\documentclass[11pt]{article}
\usepackage[hyperref, final]{acl}
\usepackage[T1]{fontenc}
\usepackage{times}
\usepackage{latexsym}
\usepackage{microtype}
\usepackage{inconsolata}
\usepackage{graphicx}
\usepackage{booktabs}
\usepackage{multirow}
\usepackage{amsmath}
\usepackage{amssymb}
\usepackage{xcolor}
\usepackage{enumitem}
\usepackage{array}
\usepackage{listings}
\usepackage{kotex}
\usepackage{caption}
\usepackage{listings}
\usepackage[most]{tcolorbox}
\tcbuselibrary{breakable}   %
\usepackage{makecell}
\usepackage{tabularx}
\title{Multi-Step Tool-Calling over Korean Open Public APIs:

A Benchmark and a Data-Synthesis Recipe}

\author{
  Dain Kim$^{*}$  \quad Eungi Cho$^{*}$  \quad Kyumin Kim$^{*}$  \quad Shinyeong Noh$^{*}$  \quad Kyuseong Lim$^{*}$  \\
  LG CNS \\
  \texttt{\{dainkim, eungizoa, kyuminkim, sy.noh, ks.lim\}@lgcns.com} \\
  \vspace{2pt}
  {\small $^{*}$Equal contribution.}
}

\begin{document}
\maketitle

\begin{abstract}

Data-sovereignty regulations increasingly require public institutions to deploy open-source, on-premise LLM agents that chain multiple tool-calls across live government APIs. However, open-source models consistently underperform in this multi-step setting, and no existing benchmark measures the gap. 
We introduce the Korean Open Public API Benchmark (\textsc{KOPA-Bench}), comprising 145 real-world tasks. 
To close this gap, we present \textsc{EDGE}, an Execution-grounded Dynamic Graph for tool-calling data synthEsis driven by live execution. 
\textsc{EDGE} builds a graph of how each tool's output can feed another's input, keeps only the links that succeed when actually called against the live APIs, and traverses these verified links to synthesize executable multi-step trajectories. Fine-tuned via GRPO on the resulting dataset, our 9B model nearly matches the untuned 27B model from the same family, improving substantially not only on \textsc{KOPA-Bench} but also on the BFCL benchmark.\footnote{Code and data are available at: {\small\url{https://github.com/dneirfi/EDGE-KOPA}}.}
\end{abstract}

\section{Introduction}
\label{sec:intro}

\begin{figure}[t]
  \centering
  \includegraphics[width=\columnwidth]{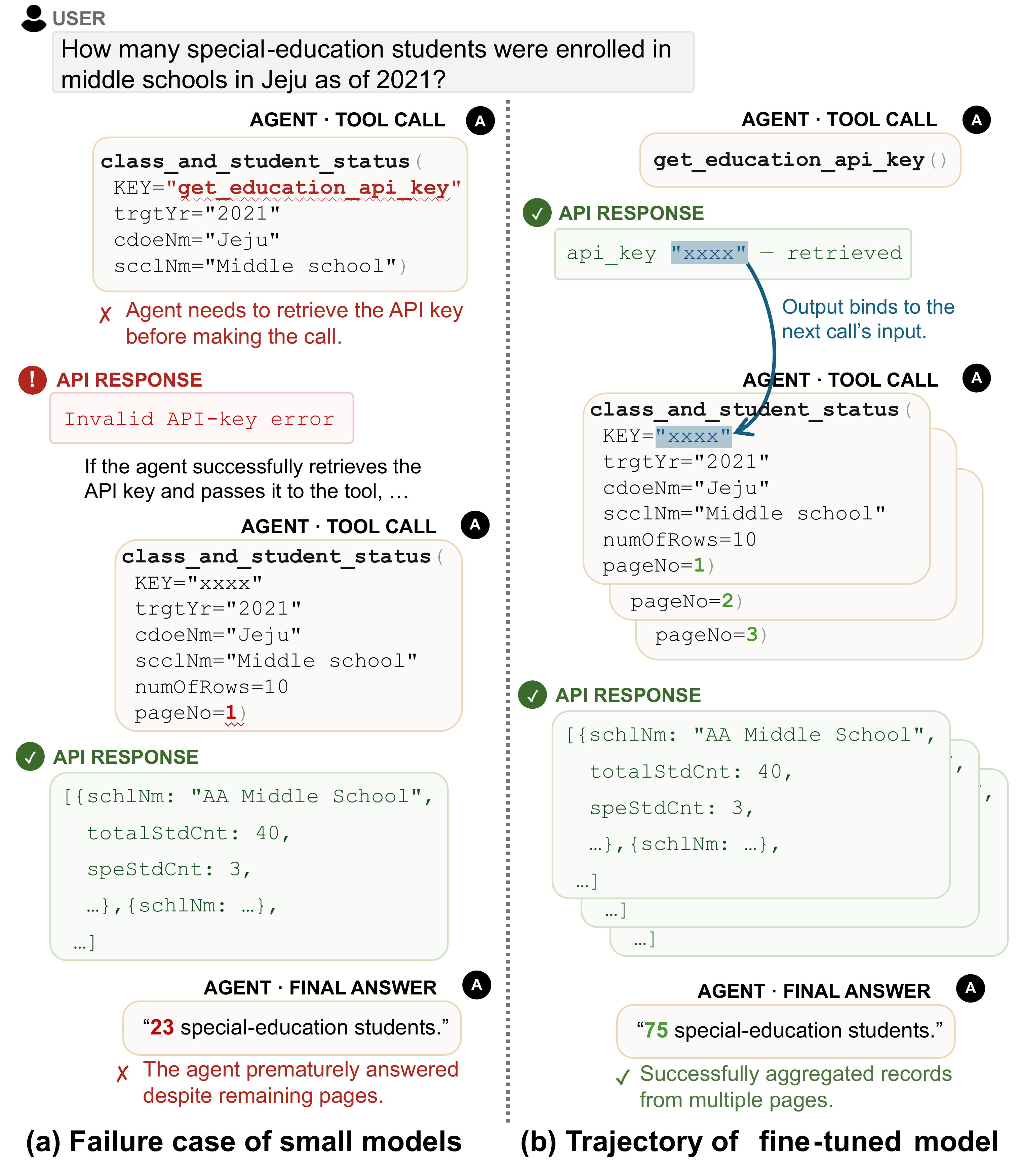}
  \caption{Comparison of tool-calling trajectories in Korean public APIs before and after fine-tuning. (a) Failure case (Before fine-tuning): The agent fails to retrieve the API key and, even after successfully retrieving a valid key, it answers incompletely based only on the first page of a paginated API response. (b) Success case (After fine-tuning): The agent  successfully chains tool outputs, queries multiple pages, and aggregates the results to deliver an accurate answer.}
  \label{fig:motivation}
\end{figure}

Public institutions are increasingly adopting agents built on large language models~\citep{Tang2025EmpoweringRA, Mohammadi2025EvaluationAB}. Data-sovereignty regulations force these agencies to deploy open-source models \citep{qwen3.5,gemma2026gemma4} on local infrastructure rather than commercial cloud services. However, these models must operate over the laws, corporate disclosures, and transportation data that the institutions provide through open APIs, a context where open-source models consistently underperform.

We study this by examining how tool-calling agents operate over live Korean public API endpoints. Under this context, we observe two distinct challenges: (1) frequent entity code-lookups that transform simple queries into dependent chains, and (2) high-cardinality responses requiring data reduction or fan-out. Open-source models consistently stumble on both, either skipping prerequisite lookups or prematurely answering from partial results when facing multi-record outputs (Figure~\ref{fig:motivation}).

We therefore construct the Korean Open Public API Benchmark (\textsc{KOPA-Bench}), 145 multi-step tasks over six domains of live Korean public APIs. Each task requires chaining dependent calls, where the output of one tool supplies the input of the next, and a lot of calls return multiple records. Improving on these tasks requires training data under the same constraints, yet existing tool-calling datasets target neither Korean public APIs, live endpoint failures, nor high-cardinality handoffs. 

We therefore propose Execution-grounded Dynamic Graph for tool-calling data synthEsis (\textsc{EDGE}), a pipeline designed around these properties. \textsc{EDGE} builds a tool dependency graph and verifies each LLM-proposed dependency against the live APIs, keeping the graph dynamic by pruning those that repeatedly fail under execution. 

From the verified graph, \textsc{EDGE} synthesizes executable trajectories by labeling each output-to-input handoff between tools with the number of distinct returned values. This lets EDGE turn high-cardinality responses into valid sequential trajectories through bounded fan-out or deterministic reduction. 

Trained on the resulting dataset, both small Qwen3.5 models improve substantially, raising \textsc{KOPA-Bench} pass@1 by $+13$pp (4B) and $+10$pp (9B), with the gains extending to the out-of-distribution BFCL benchmark~\citep{patil2025bfcl}.

Our contributions are as follows:
\begin{itemize}
\item We construct \textsc{KOPA-Bench}, a multi-step function-calling benchmark on live Korean public APIs, where each task chains dependent calls through code-based lookups and high-cardinality intermediate results.
\item We propose \textsc{EDGE}, a synthesis pipeline that verifies tool dependencies by live execution and resolves high-cardinality handoffs into sequential trajectories. GRPO training on its data substantially improves small open-source models on \textsc{KOPA-Bench} and out-of-distribution benchmarks.
\end{itemize}

\section{Benchmark Construction}
\label{sec:benchmark}
We introduce Korean Open Public API Benchmark \textsc{KOPA-Bench}, a multi-step function-calling benchmark over Korean Open Public APIs, comprising 145 tasks across 10 platforms and six domains. Each task requires chaining tool-calls, where one call's output feeds the next's input. Unlike prior benchmarks built on emulated services \citep{patil2025bfcl} or hand-authored simulations with an LLM user simulator \citep{barres2025tau2}, \textsc{KOPA-Bench} is grounded in real public APIs.

\subsection{Platform Selection}
\label{subsec:platform}
We select platforms that satisfy three criteria: (1)~\textbf{domain coverage} across traffic, finance, education, law, politics and district administration—domains where public institutions rely on open APIs; %
(2)~\textbf{API interconnectivity}, enabling chained multi-step reasoning; and (3)~\textbf{data licensing} permitting derivative works under the Korea Open Government License (KOGL). 
This yields 10 platforms spanning six domains. On average, a task requires five tool-calls (up to 14), with 59\% involving parallel execution. Per-domain statistics and the full platform list are provided in Appendix~\ref{app:stats}.

\subsection{Tool Construction}
\label{subsec:tool}
For each platform, we parse the official documentation to implement a Model Context Protocol (MCP) server that exposes typed function signatures with natural-language descriptions, allowing direct invocation via standard LLM function-calling interfaces. The server integrates an API client handling authentication, session management, error handling, and retry logic. It also manages per-task environment state, initializing it at the start of each evaluation episode to support ENVIRONMENT-based evaluation (\S\ref{subsec:eval}).  In total, this yields a tool inventory of 2{,}318 functions across the 10 platforms, each backed by a live endpoint.

\subsection{Task Generation}
\label{subsec:task}
Based on the deployed tools, domain experts design chainable tool sets for each
domain and formulate queries that necessitate multi-step tool invocations. Each
query is annotated with golden actions, the ground-truth optimal trajectory, as
well as the target response or environment state for evaluation.

We validate the annotations in two stages. First, an expert who is not among the
task authors audits every task against the tool descriptions, checking tool
selection, arguments, and tool responses. Second, we execute each golden
trajectory and give the resulting tool outputs to a frontier model (Claude
Sonnet 4.6), then compare its response against the annotated target. The audit
finds errors in 7 tasks ($4.8\%$), and, excluding cases of model failure, $80\%$
of tasks pass the execution check on the first attempt. We correct every task
flagged by either stage and repeat the check until all 145 pass.

\subsection{Evaluation Methodology}
\label{subsec:eval}
We evaluate along three dimensions---RESPONSE, ENVIRONMENT, and ACTION---corresponding to the model's final answer, the resulting server state, and the executed tool-calls.
Following $\tau^2$-Bench~\citep{barres2025tau2}, each task carries \texttt{reward\_basis} annotation that designates the subset of metrics by which it is scored. ACTION is always included, while RESPONSE and ENVIRONMENT follow the response-based and state-based paradigms of BFCL: RESPONSE applies when the task demands a definite answer, and ENVIRONMENT when it depends on the resulting system state. These complementary criteria can be applied individually or in tandem.

\paragraph{Evaluation criteria.} %
ACTION-only evaluation is insufficient (Appendix~\ref{app:action}), as a trajectory may contain the golden action yet leave the system in an incorrect state, or conversely reach the correct outcome via an alternative sequence. We therefore evaluate along two separate axes: RESPONSE and ENVIRONMENT verify the outcome, while ACTION scores the trajectory that produced it. Detailed procedures are in Appendix~\ref{app:eval}, and evaluation prompts in Appendix~\ref{app:eval-prompts}.

\section{Methods}
\label{sec:datagen}

\begin{figure*}[t]
  \centering
  \includegraphics[width=\textwidth]{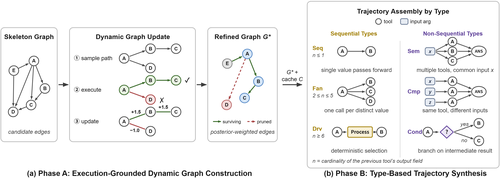}
  \caption{\textbf{Overview of EDGE.}
  \textbf{(a)}~Phase~A constructs an execution-grounded dynamic graph: from
  a skeleton of LLM-proposed candidate edges, it revises both edge posteriors
  and its topology using live execution evidence, yielding a refined
  graph $G^\star$. \textbf{(b)}~Phase~B assembles trajectories over
  $G^\star$ by junction type and generates a natural-language query for
  each, producing the training dataset.}
  \label{fig:main_pipeline}
\end{figure*}

We introduce Execution-grounded Dynamic Graph for tool-calling data synthEsis (\textsc{EDGE}), a two-phase framework that synthesizes multi-step tool-calling data from live APIs (Figure~\ref{fig:main_pipeline}). Phase~A (\S\ref{sec:edge-phasea}) builds a dependency graph, keeping only edges that execute successfully. Phase~B (\S\ref{sec:method:trace}) assembles trajectories over the graph, typing each junction by response cardinality, and generates a Korean query and answer for each.

\subsection{Problem Setting}
\label{sec:problem_setting}
We construct a tool inventory $\mathcal{V}=\{v_1,\dots,v_N\}$ comprising $N{=}2{,}318$ tools across six domains. Each tool $v \in \mathcal{V}$ exposes a live endpoint and a JSON-typed signature $\sigma_v=(\mathrm{desc}_v,\mathcal{I}_v,\mathcal{O}_v)$, which consists of a Korean description, an input-parameter schema and an output-field schema. Invoking $v$ returns a set of records $R_v$ whose cardinality ranges from zero to hundreds of thousands. Thus, a chain $u\!\to\!v$ (where $u, v \in \mathcal{V}$) has no single output to bind, a challenge that we address in \S\ref{sec:method:trace}.

\subsection{Phase A: Execution-Grounded Dynamic Graph Construction}
\label{sec:edge-phasea}

\paragraph{Skeleton graph construction.}
We take the tool inventory $\mathcal{V}$ in \S\ref{sec:problem_setting} as
the nodes of a directed dependency graph $G=(\mathcal{V},\mathcal{E})$ (Figure~\ref{fig:main_pipeline}a), 
where an edge $e=(u\to v)$ asserts that an output field of $u$ binds to an input parameter of $v$. Since scoring
all $\mathcal{O}(|\mathcal{V}|^2)$ pairs with an LLM is infeasible, a dense retriever over signature embeddings restricts each source $u$ to its top-$K$ neighbors: 15 from the same domain and 10 from other domains, where bindings are rarer.

For each candidate edge, a single LLM call returns a feasibility score $s_e\in[0,1]$, a binding set $B_e$ that maps bound parameters of $v$ to the supplying output of $u$, and default candidates $D_e$ for %
left unbound. These compile into a source plan over every parameter $p\in\mathcal{I}_v$:
\begin{equation}
\Pi_e(p)\;=\;
\begin{cases}
\textsc{Upstream}\,\langle B_e(p)\rangle & p\in\mathrm{dom}(B_e),\\
\textsc{Default}\,\langle d_p\rangle & p~\text{generic},\\
\textsc{Clarify}\,\langle D_e(p)\rangle & \text{otherwise,}
\end{cases}
\label{eq:sourceplan}
\end{equation}
where $\mathrm{dom}(B_e)$ denotes the set of parameters bound by $B_e$.
 \textsc{Upstream} binds $p$ to its supplying output $B_e(p)$ from the upstream tool $u$, \textsc{Default} is a static lookup of generic parameters (API keys, etc.), and \textsc{Clarify} tries $D_e(p)$ in the LLM's proposals. We admit an edge only if its feasibility score meets the threshold.
 Each admitted edge receives a feasibility-anchored Beta prior $\alpha_e=1+\kappa s_e$, $\beta_e=1+\kappa(1-s_e)$, where pseudo-counts $\alpha_e$ and $\beta_e$ favor success and failure in proportion to $s_e$, with $\kappa$ controlling the prior's strength.

\paragraph{Execution-grounded graph update.}
Each iteration samples a batch of paths, executes them against live APIs, and revises the edge posteriors and the graph from the outcomes (Figure~\ref{fig:main_pipeline}a). The graph is \emph{dynamic}, with both its posteriors and its topology changing as execution evidence accumulates. 

We select edges by Thompson sampling~\citep{agrawal2013thompson}, a posterior-sampling approach that balances \emph{exploitation} and \emph{exploration}. We model the success rate of each edge $e$ as $\theta_e$ with a Beta posterior $\mathrm{Beta}(\alpha_e,\beta_e)$, whose pseudo-counts $\alpha_e$ and $\beta_e$ accumulate the successful and failed executions of the edge. From the current node, we sample $\tilde\theta_e\sim\mathrm{Beta}(\alpha_e,\beta_e)$ per outgoing edge and follow the largest, with an $\varepsilon$-greedy fallback to a random edge:
\begin{equation}
e^\star =
\begin{cases}
\arg\max_e \tilde\theta_e & \text{w.p. } 1-\varepsilon,\\[2pt]
\text{a random outgoing edge} & \text{w.p. } \varepsilon.
\end{cases}
\label{eq:thompson}
\end{equation}
Because each $\tilde\theta_e$ comes from the full posterior rather than its mean, an edge with a high posterior mean usually yields a large sample and is exploited often, while an edge with few trials has a wide posterior that occasionally yields a large sample and is therefore still explored. 

The $\varepsilon$-greedy term sets a floor on exploration, so cold-start edges that Thompson sampling alone might never select are eventually tried. To run a path, the first tool receives no upstream output, so its arguments come from a single LLM call at execution time (with a rule-based fallback), and every subsequent argument is filled by the source plan $\Pi_e$ of Eq.~(\ref{eq:sourceplan}). Each outcome is \emph{structural} when the binding itself fails or \emph{environmental} when a transient server-side error occurs, and structural failures raise $\beta_e$ far more than environmental ones, because random server errors would otherwise weaken and eventually prune correctly bound edges. 

An edge is pruned when the posterior probability that its success rate exceeds $\tau_{\mathrm{prune}}$ falls below $\varepsilon_{\mathrm{prune}}$, after at least $n_{\min}$ trials. Iterating this loop to convergence yields the refined graph $G^\star$, which retains only the admitted edges that survive pruning. The full update scheme, hyperparameters, and supporting analyses are provided in Appendix~\ref{app:app-egdg}. Phase~B (\S\ref{sec:method:trace}) consumes $G^\star$ and a per-edge cache $\mathcal{C}$ of observed (input, first-record) pairs to ground query synthesis.

\begin{table*}[t]
    \centering
    \small
    \setlength{\tabcolsep}{6pt}
    \renewcommand{\arraystretch}{1.1}
    \begin{tabular}{l ccc ccc}
        \toprule
        & \multicolumn{3}{c}{\textbf{\textsc{KOPA-Bench}}}
        & \multicolumn{3}{c}{\textbf{BFCL}} \\
        \cmidrule(lr){2-4} \cmidrule(lr){5-7}
        \textbf{Model}
        & \textbf{pass@1} & \textbf{pass@4} & \textbf{Action}
        & \textbf{\makecell{Multi-\\Turn}}
        & \textbf{\makecell{Single\\(non-live)}}
        & \textbf{\makecell{Single\\(live)}} \\
        \midrule
        \multicolumn{7}{l}{\textit{Proprietary}} \\
        \quad Claude Sonnet 4.6~\citep{anthropic2026sonnet46}
            & 0.4655 & 0.8207 & 0.4269
            & 60.13 & 84.98 & 76.83 \\
        \quad GPT-5.1~\citep{openai2025gpt51}
            & 0.3706 & 0.4690 & 0.5881
            & 38.12 & 82.96 & 65.21 \\
        \addlinespace[2pt]
        \multicolumn{7}{l}{\textit{Large open-source}} \\
        \quad Qwen3.5-27B
            & 0.4482 & 0.5655 & 0.4213
            & 65.38 & 89.65 & 83.35 \\
        \quad gemma-4-26B-A4B-it
            & 0.3534 & 0.4690 & 0.5061
            & 55.38 & 81.23 & 80.16 \\
        \quad EXAONE-4.5-33B~\citep{choi2026exaone45technicalreport}
            & 0.2586 & 0.3862 & 0.2670
            & 53.00 & 87.52 & 80.16 \\
        \addlinespace[2pt]
        \multicolumn{7}{l}{\textit{Small open-source}} \\
        \quad Qwen3.5-9B
            & 0.3275 & 0.4690 & 0.3407
            & 52.25 & 82.40 & 78.09 \\
        \quad gemma-4-E4B-it
            & 0.1862 & 0.2207 & 0.4069
            & 20.75 & 84.79 & 63.43 \\
        \quad Qwen3.5-4B
            & 0.1758 & 0.3103 & 0.2140
            & 49.08 & 79.81 & 78.02 \\
        \quad Ministral-3-8B-Instruct-2512~\citep{liu2026ministral}
            & 0.1299 & 0.2308 & 0.3797
            & 24.50 & 82.38 & 74.54 \\
        \midrule
        \quad \textbf{Qwen3.5-4B (Ours)}
            & \textbf{0.3094} & \textbf{0.4690} & \textbf{0.3462}
            & \textbf{53.12} & \textbf{79.96} & \textbf{76.61} \\
        \quad \textbf{Qwen3.5-9B (Ours)}
            & \textbf{0.4310} & \textbf{0.5517} & \textbf{0.3655}
            & \textbf{58.12} & \textbf{87.65} & \textbf{81.57} \\
        \bottomrule
    \end{tabular}
    \caption{Results on \textsc{KOPA-Bench} and BFCL. \textbf{\textsc{KOPA-Bench}} ---
        \textbf{pass@$k$}: fraction of the $145$ tasks solved by at least one
        of $k$ rollouts, so \textbf{pass@1} is the fraction solved on a single
        attempt (averaged over the four rollouts) and \textbf{pass@4} the
        fraction solved by at least one of the four; \textbf{Action}: tool-call accuracy over golden
        actions. \textbf{BFCL} scores are computed via the official
        evaluation repository. The bottom block reports our fine-tuned
        models.}
    \label{tab:main}
\end{table*}

\subsection{Phase B: Type-Based Trajectory Synthesis}
\label{sec:method:trace}

\paragraph{Sequential trajectories.}
Given the validated edge graph (\S\ref{sec:edge-phasea}), we synthesize a sequential trajectory as a chain of tool-calls $(T_1,\ldots,T_m)$. The chain starts from an initial tool-call and follows field-level edges in the graph. 
For a junction $j_i = (T_i.f_i \rightarrow T_{i+1}.a_i)$, values returned in the output field $f_i$ of $T_i$ are used to instantiate the target argument $a_i$ of $T_{i+1}$. 
The required arguments not determined by a junction are filled from the validated cache $\mathcal{C}$. 
A key challenge with Korean Open Public APIs is that tools often return more than thousands of records.
At this scale, a junction no longer specifies a unique downstream binding. Instead, it creates many possible continuations that can make the synthesized trajectory ambiguous, invalid, or unanswerable if left unresolved. Therefore, we introduce a junction typing scheme based on the cardinality of the connecting field.

With fan-out budget $\phi=5$, we assign \emph{Pure Sequential} (\textsc{Seq}) if $n_i=1$, \emph{Fan-out} (\textsc{Fan}) if $2 \le n_i \le \phi$, and \emph{Derived} (\textsc{Drv}) if $n_i>\phi$. 
\textsc{Seq} passes the single value
to the next tool, while \textsc{Fan} issues one downstream call for each
value in $U_i$. 
For \textsc{Drv}, we insert an internal process node that selects a bounded subset before proceeding. 
We use $\max$, $\min$, ${\geq}$, and
${\leq}$ for numeric fields, equality for \texttt{enum} fields,
\texttt{date-after} for date fields, and \texttt{most-frequent} as a
type-agnostic operator. For threshold and equality filters, the pivot value is
chosen from observed field values so that the filtered set contains at most
$\phi$ distinct values.

The structural pattern of a sequential trajectory is the ordered composition of its junction types, e.g., \textsc{Seq}{+}\textsc{Fan}, which determines the query skeleton used later. 
While pipelines that assume single-value flows typically discard or mis-bind many-record responses, cardinality typing safely consumes them via bounded enumeration (\textsc{Fan}) or deterministic reduction (\textsc{Drv}).

\paragraph{Non-sequential trajectories.}
We additionally synthesize three non-sequential types from explicit templates, retaining only successfully executed paths.
\emph{Semantic-parallel} (\textsc{Sem}) calls independent tools with shared arguments and combines their outputs.
\emph{Comparison} (\textsc{Cmp}) calls the same tool with different arguments and compares the results.
\emph{Conditional} (\textsc{Cond}) evaluates the condition over an intermediate result and proceeds along the branch that outcome specifies.
These templates cover parallel, comparative, and conditional patterns that cannot be represented as a single linear chain.

\paragraph{Query generation and validation.}
For each trajectory, an LLM query generator produces a Korean natural-language query and derives the answer from the cached execution results, guided by type-specific prompt templates that describe the trajectory structure. The resulting query, answer, and execution trace then pass through a three-stage validation pipeline. Throughout, only open-source model outputs become training labels, while proprietary models serve solely for verification. Appendix~\ref{app:filter} details the pipeline with representative anomaly examples; all prompt templates are in Appendix~\ref{app:method-prompts}.

\section{Experiments}
\label{sec:experiments}

We investigate whether training on EDGE dataset improves multi-step function-calling over Korean public APIs and whether the gains generalize beyond our training distribution. We evaluate on \textsc{KOPA-Bench} to measure Korean public API tool use, and on BFCL to measure out-of-distribution performance. Appendix~\ref{app:contamination} verifies that \textsc{KOPA-Bench} is not
contaminated by the training dataset, and additionally reports performance on
platforms withheld from synthesis (Appendix~\ref{app:heldout}) and the overlap of
dependency edges between the two corpora (Appendix~\ref{app:edgeoverlap}).

\subsection{Setup}
We fine-tune Qwen3.5-4B and Qwen3.5-9B with GRPO on the $1{,}781$ training data of \S\ref{sec:datagen} (detailed in Appendix~\ref{app:corpus}), employing a binary reward $r\in\{0,1\}$ over the RESPONSE and ENVIRONMENT dimensions. Training with \texttt{verl}~\citep{sheng2024hybridflow} on $8{\times}$H100
GPUs, full details in Appendix~\ref{app:training}. We sample four independent rollouts per task and report both pass@1 and pass@4. To assess the statistical reliability of the pass@1 estimates, we additionally re-evaluate all models over eight independent random seeds and report $95\%$ confidence intervals in Appendix~\ref{app:ci}.

\subsection{Main results}
\paragraph{\textsc{KOPA-Bench}.}
As shown in Table~\ref{tab:main}, both fine-tuned models achieve substantial
gains over their base counterparts. Our 4B model raises pass@1 from $0.18$ to $0.31$
($+13$pp) and the 9B model from $0.33$ to $0.43$ ($+10$pp), approaching Qwen3.5-27B ($0.45$) at one-third of its parameters. The $95\%$ confidence
intervals estimated over eight independent seeds do not overlap between our
models and their base counterparts, so the gains are statistically significant
(Appendix~\ref{app:ci}). They also extend beyond the platforms seen during
synthesis: on the 31 tasks grounded in platforms withheld from synthesis
entirely, the 4B model improves by $+22.6$pp in pass@4, exceeding its $+15.9$pp pass@4 gain over
the full benchmark (Appendix~\ref{app:heldout}).

\paragraph{Other benchmark.}
Training on the EDGE corpus does not come at the cost of out-of-distribution
performance. On the BFCL benchmark, both fine-tuned models on aggregate improve over their base counterparts. Multi-turn improves the most (+4.04pp for 4B, +5.87pp for 9B), consistently exceeding the single-turn gains. Despite our training distribution being predominantly single-turn, the multi-turn generalization likely stems from the structural alignment between multi-step tool-call trajectories and multi-turn conversational structures.

\subsection{Ablation Study}
\label{sec:ablation}

\begin{table}[t]
    \centering
    \small
    \setlength{\tabcolsep}{5pt}
    \renewcommand{\arraystretch}{1.15}
    \begin{tabular}{l ccc}
        \toprule
        \textbf{Training objective}
        & \textbf{pass@1} & \textbf{pass@4} & \textbf{Action} \\
        \midrule
        Qwen3.5-4B (base)
            & 0.1758 & 0.3103 & 0.2140 \\
        \addlinespace[2pt]
        \quad SFT (Ours)
            & 0.2724 & 0.3586 & 0.3048 \\
        \quad GRPO (Ours)
            & \textbf{0.3094} & \textbf{0.4690} & \textbf{0.3462} \\
        \bottomrule
    \end{tabular}
    \caption{Training objective ablation, evaluated on \textsc{KOPA-Bench} with four rollouts per task; metrics are defined as in Table~\ref{tab:main}. SFT and GRPO consume the identical $1{,}781$ EDGE tasks and share the same base checkpoint.}
    \label{tab:objective-ablation}
\end{table}

\paragraph{Effect of the training objective.}
The gains in Table~\ref{tab:main} could in principle come from the GRPO
objective rather than from the EDGE corpus itself. To isolate the objective's
contribution, we supervise-fine-tune the same base checkpoint on the same
$1{,}781$ tasks. We generate trajectories with Qwen3.5-397B-A17B under the
environment used for GRPO rollouts (identical tools, system prompt, and runtime
context) and train on the verified ones.

As shown in Table~\ref{tab:objective-ablation}, SFT alone already raises pass@1
by $+9.7$pp and Action by $+9.0$pp over the base model, so most of the
improvement is attributable to the corpus rather than to the objective. GRPO
then adds a further consistent gain across all three metrics, most visibly in
pass@4 ($0.3586\rightarrow0.4690$, $+11.0$pp). Since both methods consume the identical task
set, this gap reflects how much signal each extracts per task rather than data
quantity. SFT imitates a single verified reference trajectory and is
upper-bounded by the teacher, whereas GRPO samples multiple rollouts per prompt
and learns from their relative rewards. Where one query admits several valid
trajectories and intermediate failures are common
(Appendix~\ref{app:action}), this exploration extracts more signal than
single-reference imitation.

\paragraph{Effect of trajectory diversification.}
To assess whether the structural diversity introduced by our synthesis pipeline drives the observed gains, we split the corpus into purely sequential, purely parallel, and \textsc{Mixed} trajectories---the last combining both structures within a single trajectory and forming the largest group (36.7\%, Table~\ref{tab:type-distribution})---and train on three subsets: \emph{pure-sequential}, \emph{parallel\,+\,\textsc{Mixed}}, and \emph{full}. Note that \emph{parallel\,+\,\textsc{Mixed}} removes only purely sequential trajectories, since the retained \textsc{Mixed} ones still carry sequential sub-chains. As shown in Table~\ref{tab:traj-ablation}, \emph{pure-sequential} lags behind (0.2327 pass@1) because it never observes parallel composition, whereas \emph{parallel\,+\,\textsc{Mixed}} (0.3080) is exposed to both and approaches \emph{full} (0.3094). The two separate at pass@4, where \emph{full} scores 0.4690 and \emph{parallel\,+\,\textsc{Mixed}} 0.4000. \emph{full} trails \emph{parallel\,+\,\textsc{Mixed}} on \textsc{Action}
(0.3462 vs.\ 0.4070), but the gap is one of task composition, not trajectory
quality. \textsc{Action} is an efficiency-weighted score averaged only over the
tasks a model solves (Appendix~\ref{app:action}), and the tasks that
\emph{full} alone solves are longer-horizon ones, on which even correct
trajectories run long and score lower on efficiency. Since adding purely sequential data causes no pass@1 regression while improving pass@4, we adopt the full mixture for our final model. The two families are complementary: sequential trajectories teach precise output-to-input handoffs across junctions, while parallel ones infuse comparative and conditional patterns that a single sequential chain cannot express.

\begin{figure}[t]
  \centering
  \includegraphics[width=\columnwidth]{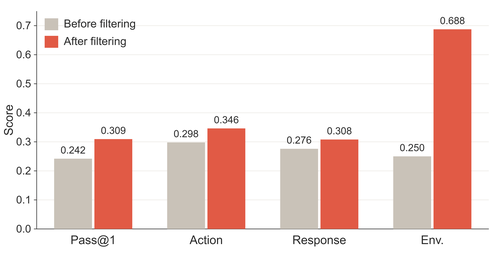}
  \caption{Effect of trajectory-level filtering on the dataset.
  Filtering improves every metric.}
  \label{fig:filter-ablation}
\end{figure}

\paragraph{Effect of data filtering.}
We investigate the effect of filtering the synthesized training data. LLM synthesis introduces anomalies such as malformed structures and stale labels from time-varying APIs (Appendix~\ref{app:filter}). Removing these tasks raises pass@1 from $0.242$ to $0.309$ ($+6.7$pp), with consistent gains across all metrics (Figure~\ref{fig:filter-ablation}).

\paragraph{Effect of execution-grounded dynamic graph.}

\begin{table}[t]
    \centering
    \small
    \setlength{\tabcolsep}{5pt}
    \renewcommand{\arraystretch}{1.15}
    \begin{tabular}{l ccc}
        \toprule
        \textbf{Training dataset}
        & \textbf{pass@1} & \textbf{pass@4} & \textbf{Action} \\
        \midrule
        Qwen3.5-4B (base)
            & 0.1758 & 0.3103 & 0.2140 \\
        \addlinespace[2pt]
        \quad pure-sequential
            & 0.2327 & 0.3724 & 0.3648 \\
        \quad parallel\,+\,\textsc{Mixed}
            & 0.3080 & 0.4000 & \textbf{0.4070} \\
        \quad full
            & \textbf{0.3094} & \textbf{0.4690} & 0.3462 \\
        \bottomrule
    \end{tabular}
    \caption{Trajectory composition ablation, evaluated on \textsc{KOPA-Bench} with four
     rollouts per task; metrics are defined as in Table~\ref{tab:main}. \textsc{Mixed}
     trajectories are retained in both \emph{parallel\,+\,\textsc{Mixed}} and \emph{full};
     \emph{pure-sequential} contains only purely sequential trajectories. \emph{full} uses
     the complete filtered dataset and corresponds to our final model, Qwen3.5-4B (Ours).}
     
    \label{tab:traj-ablation}
\end{table}

\begin{figure}[t]
  \centering
  \includegraphics[width=\columnwidth]{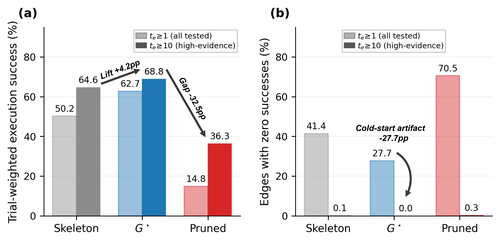}
  \caption{\textbf{Execution-grounded refinement improves the graph and
  prunes the right edges.} Edges are grouped as the pre-refinement
  Skeleton, the surviving graph $G^\star$, and the
  Pruned set; %
  light and dark bars count all tested edges ($t_e\!\ge\!1$) and high-evidence
edges ($t_e \ge 10$); the $t_e \ge 10$ bars are not comparable across groups,
since failing edges are pruned before accumulating trials.}
  \label{Fig:edge-precision}
\end{figure}

Phase~A first builds a skeleton graph from LLM feasibility scores, then prunes edges whose posteriors fall too low (\S\ref{sec:edge-phasea}). We ask whether this pruning truly separates executable from non-executable edges, rather than discarding them at random. Figure~\ref{Fig:edge-precision} compares the execution success rate of three
edge sets, namely the LLM-judged skeleton, our converged $G^\star$, and the
pruned set removed during the loop. The skeleton serves as a controlled proxy
for how prior synthesis pipelines establish links between
calls~\citep{yin-etal-2025-magnet, liu2025toolace, prabhakar2026apigenmt}. It
fixes each dependency from signature matching and a single LLM judgment, and it
never checks that dependency against a live endpoint. It shares only this
property with the prior pipelines by construction and is not an end-to-end
reimplementation of any of them, which lets the comparison isolate the effect of
execution grounding while holding the candidate edge set fixed.

The results show this separation clearly. Only $50.2\%$ of the skeleton edges
actually execute, whereas $G^\star$ reaches $62.7\%$ ($+12.5$pp). Since both
start from the same LLM-proposed edges, this gain comes entirely from execution,
because the loop removes edges the LLM judged plausible but that fail in
practice. The pruned edges, in turn, execute only $14.8\%$ of the time, far
below $G^\star$. If pruning were random, the two sets would execute at similar
rates. Instead, the loop removes exactly the edges that live APIs reject, which
any pipeline that fixes dependencies without executing them cannot detect.
Figure~\ref{Fig:edge-precision}b confirms that the loop prunes dead edges rather
than promising ones. Among the pruned edges, $70.5\%$ never succeed in any
trial, against $27.7\%$ of those retained in $G^\star$. The residual
zero-success edges in $G^\star$ are concentrated among edges tried only a few
times, so they reflect under-exploration rather than demonstrated failure.

\section{Conclusion}
\label{sec:conclusion}

In this work, we address multi-step tool-calling over live Korean public APIs by introducing \textsc{KOPA-Bench}, which reveals performance limitations in existing LLMs. To bridge this gap, we present \textsc{EDGE}, an execution-grounded framework that synthesizes verified multi-step trajectories. 
Our findings demonstrate that training small open-source models with GRPO on the resulting dataset yields substantial gains, nearly matching much larger architectures and generalizing to BFCL. Thus, \textsc{KOPA-Bench} and \textsc{EDGE} offer a scalable foundation for deploying reliable LLM agents in public applications.

\section*{Limitations}
\label{sec:limitations}

\paragraph{Reliance on live endpoints.} Because the benchmark and the synthesized dataset are grounded in live APIs, both are subject to endpoint drift: schemas, availability, and returned records can change over time. We mitigate this by filtering tasks tied to real-time data sources (Appendix~\ref{app:filter}) and by evaluating environment state through hash comparison, but exact reproduction still depends on the stability of public services we do not control.

\paragraph{Beyond Korean public APIs.} Both \textsc{KOPA-Bench} and the \textsc{EDGE} pipeline are built on Korean public-sector APIs across six domains. This isolates the multi-step, high-cardinality setting we target, but leaves open how far our findings transfer to other languages, to commercial or private APIs, and to administrative systems in other countries. We leave extending \textsc{EDGE} to these broader settings to future work.

\paragraph{Scope of comparison.}Our ablations compare \textsc{EDGE} against controlled variants of itself, including a schema-only skeleton that approximates settings in which tool dependencies are specified but not executed. This design isolates the contribution of execution grounding while holding the surrounding pipeline fixed. Running an existing synthesis system end to end on our tool inventory and training on its output remains outside the scope of the present study, so system-level comparisons with prior synthesis pipelines are left open.

\section*{Acknowledgements}

This work was conducted as part of the Sovereign AI Foundation Model Project (GPU Track), organized by the Ministry of Science and ICT (MSIT), South Korea (PJT-26-010016). 

We thank LG AI Research for valuable discussions and feedback on this work.

\bibliography{main}
\appendix
\section{Related Work}
\label{sec:related}

\paragraph{Function-calling benchmarks.}
Function-calling evaluation has been driven by foundational benchmarks such as API-Bank~\citep{li-etal-2023-api} and BFCL. To broaden scope, ToolBench~\citep{qin2024toolllm} scales to thousands of real-world REST APIs, StableToolBench~\citep{guo2024stabletoolbench} adds simulated environments to address the instability of live APIs, and $\tau^2$-Bench~\citep{barres2025tau2} scores agents against the resulting environment state through a user simulator. For Korean, FunctionChat-Bench~\citep{lee2024functionchat} evaluates general-purpose tools and OrchestrationBench~\citep{ahn2026orchestrationbench} evaluates tool coordination that requires structured planning.
None, however, targets public-sector APIs; we introduce a benchmark over live Korean public APIs where solving a query requires chaining interdependent calls over many record responses.

\paragraph{Function-calling data synthesis.}
Since real function-calling data are scarce, a common approach is to synthesize them with LLMs. To keep synthesized calls valid, ToolACE~\citep{liu2025toolace} adds a verification stage, and AWM~\citep{maekawa2026towards} generates calls inside simulated API environments that return consistent responses. For multi-step tasks Magnet~\citep{yin-etal-2025-magnet}, BUTTON~\citep{chen2025facilitating}, and APIGen-MT~\citep{prabhakar2026apigenmt} build such call chains by linking functions through their signatures or LLM-proposed plans. In all of them, links between calls are fixed without checking them against the live APIs, and each call is assumed to return a single result, leaving unaddressed the unstable endpoints and multi-record responses of real public APIs.

\section{Evaluation Details}
\label{app:eval}

\subsection{Task Example}
\label{app:task-example}

\begin{figure}[t]
  \centering
  \includegraphics[width=\columnwidth]{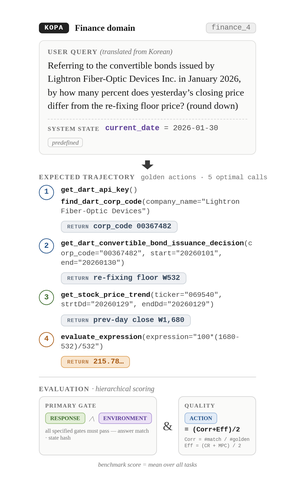}
  \caption{A representative task from the Finance domain of \textsc{KOPA-Bench}. Each task pairs a natural-language query with system state injected at runtime. The model calls the provided tools to solve the query and output a final answer, which is evaluated against the annotated ground truth (golden actions, expected response, and resulting environment state).}
  \label{fig:task_example}
\end{figure}

We illustrate the benchmark with a representative task drawn from the finance domain (Figure~\ref{fig:task_example}). The task requires the model to compute how far a stock's latest closing price sits above the refixing floor of a convertible bond, given the current date. The user query is \textit{``Referring to the convertible bonds issued by Lightron Fiber-optic Devices in January 2026, calculate by what percentage yesterday's closing price differs from the refixing floor price.''} Figure~\ref{fig:task_example} shows the full task definition and the evaluation workflow.

To solve this task, the model must resolve the company's DART corporate code, retrieve its January 2026 convertible-bond issuance record to obtain the refixing floor price, look up the prior day's closing price, and finally compute the percentage gap. The expected trajectory is:

\begin{lstlisting}[breaklines=true,basicstyle=\footnotesize\ttfamily,columns=fullflexible]
1. find_dart_corp_code(company_name="Lightron")
   -> corp code "00367482"
2. get_dart_convertible_bond_issuance_decision(
     corp_code="00367482", start="20260101",
     end="20260130")
   -> refixing floor 532
3. get_stock_price_trend(ticker="069540",
     strtDd="20260129", endDd="20260129")
   -> closing price 1680
4. evaluate_expression("100 * (1680 - 532) / 532")
   -> 215.78...
\end{lstlisting}

The system state includes context injected into the agent's system prompt at runtime. \texttt{PredefinedSystemState} entries supply static values (e.g., \texttt{current\_date: 2026-01-30}), while \texttt{ActionSystemState} entries are resolved dynamically by executing the specified function at the start of each episode (e.g., retrieving the API key via \texttt{get\_dart\_api\_key()}).

\subsection{RESPONSE Evaluation}

\label{app:response}

Response evaluation verifies whether a model's final answer matches the ground truth. We support three answer types, each appending a format-specifying suffix to the system prompt.

\textbf{String type.} The model emits its answer as \texttt{ANSWER:\{answer\}}. We extract it via pattern matching, normalize it by removing whitespace and special characters, and compare it against all acceptable values. Unmatched cases fall back to LLM-as-a-Judge.

\textbf{Number type.} The model encloses its answer in \verb|\boxed{}| as a \texttt{sympy}-parseable number or expression (e.g., \verb|\boxed{42}|). We normalize the extracted content through comma removal and float conversion, evaluating non-numeric strings symbolically with \texttt{sympy}, then compare to the ground truth under approximate equality. Failed cases fall back to LLM-as-a-Judge.

\textbf{Judge type.} For open-ended tasks such as those in the Law domain, the model answers in the same format, and we rely solely on LLM-as-a-Judge without pattern-based extraction.

\subsection{ENVIRONMENT Evaluation}

\label{app:environment}

We use \textit{golden} to denote the annotated ground-truth value throughout the remainder of the paper. ENVIRONMENT evaluation verifies that the system state resulting from the model's tool execution matches the golden state. Each task runs in a freshly launched MCP server instance, isolating its state from other tasks so that evaluation is unaffected by residual state. At the end of each episode, we retrieve the final state and compare it against the golden state annotation via SHA-256 hash comparison, following \citet{yao2024tau}.
\subsection{ACTION Evaluation}

\label{app:action}

ACTION evaluation measures both the correctness and efficiency of the model's tool-call trajectory.
\paragraph{Motivation.} Prior function-calling benchmarks such as BFCL evaluate models primarily through action correctness, verifying whether predicted tool-calls match the golden actions. Correctness alone is insufficient for two reasons. First, a golden action's presence does not guarantee that \textit{only} correct actions were taken. Given the golden action \texttt{fillTank(amount=20)}, the trajectory \texttt{[fillTank(amount=20), fillTank(amount=10)]} contains it yet reaches an incorrect final state. Second, one outcome is often reachable through multiple valid sequences. \texttt{[fillTank(amount=10), fillTank(amount=10)]} produces the same state as \texttt{fillTank(amount=20)} but is penalized under strict matching. We therefore use ACTION not as a primary success criterion but as a quality measure gated by outcome verification (\S\ref{subsec:eval}), decomposed into correctness and efficiency.

\paragraph{Tool-call correctness.} For each ground-truth action, we search the entire predicted trajectory for a matching call, without requiring that calls appear in the annotated order. A match requires that the function name is identical, all required parameters are present, no unexpected parameters are included, parameter types conform to the schema under BFCL-style AST type checking, and parameter values match the ground truth for the arguments listed in \texttt{value\_compare\_args}. The correctness score $\mathrm{Corr}$ is the fraction of golden actions successfully matched.

\paragraph{Tool-call efficiency.}
Efficiency combines a duplicate-call ratio ($\mathrm{CR}$) and minimal-path coverage ($\mathrm{MPC}$):
\begin{equation}
\resizebox{\columnwidth}{!}{%
$\displaystyle
\mathrm{CR} = 1 - \frac{N_{\mathrm{dup}}}{N_{\mathrm{total}}},
\quad
\mathrm{MPC} = \frac{N_{\mathrm{opt}}}{N_{\mathrm{total}}},
\quad
\mathrm{Eff} = \frac{\mathrm{CR} + \mathrm{MPC}}{2}
$%
}
\end{equation}
where $N_{\mathrm{dup}}$ is the number of redundant duplicate calls, $N_{\mathrm{total}}$ the total number of calls, and $N_{\mathrm{opt}}$ the annotated optimal call count. Each task specifies a \texttt{max\_allowed\_calls} budget; if $N_{\mathrm{total}}$ exceeds it, $\mathrm{Eff}$ is set to $0$. The final ACTION score is the mean of the two, $(\mathrm{Corr} + \mathrm{Eff})/2$.

\begin{table}[t]
\centering
\small
\setlength{\tabcolsep}{4pt}
\begin{tabular}{lcc}
\toprule
\textbf{Model} & \textbf{pass@1} & \textbf{CI width} \\
\midrule
Qwen3.5-27B & 0.4681 {\scriptsize$\pm$ 0.0146} & 0.0292 \\
gemma-4-26B-A4B-it & 0.3672 {\scriptsize$\pm$ 0.0151} & 0.0301 \\
Qwen3.5-9B & 0.3526 {\scriptsize$\pm$ 0.0249} & 0.0498 \\
EXAONE-4.5-33B & 0.2224 {\scriptsize$\pm$ 0.0122} & 0.0244 \\
gemma-4-E4B-it & 0.1845 {\scriptsize$\pm$ 0.0110} & 0.0220 \\
Qwen3.5-4B & 0.1603 {\scriptsize$\pm$ 0.0204} & 0.0407 \\
Ministral-3-8B-Instruct-2512 & 0.1418 {\scriptsize$\pm$ 0.0133} & 0.0265 \\
\midrule
Qwen3.5-4B (Ours) & \textbf{0.2931} {\scriptsize$\pm$ 0.0169} & 0.0338 \\
Qwen3.5-9B (Ours) & \textbf{0.4138} {\scriptsize$\pm$ 0.0098} & 0.0195 \\
\bottomrule
\end{tabular}
\caption{Pass@1 on \textsc{KOPA-Bench}, reported as mean $\pm$ 95\% confidence-interval half-width, computed over eight independent random seeds using the Student's $t$-distribution ($\mathrm{df}=7$); the last column gives the full CI width. These eight-seed estimates complement the four-run averages reported in Table~\ref{tab:main}.}
\label{tab:ci}
\end{table}

\section{Statistical Reliability of Main Results}
\label{app:ci}
 
Given the modest size of \textsc{KOPA-Bench} (145 tasks), point estimates alone could be misleading. To assess whether the improvements reported in Table~\ref{tab:main} are statistically meaningful, we re-evaluate every model over $n{=}8$ independent random seeds and report 95\% confidence intervals for pass@1.
Note that Table~\ref{tab:main} in the main text reports the original estimates averaged over four runs, whereas this appendix reports the revised eight-seed estimates; the two are consistent, and all conclusions in \S\ref{sec:experiments} hold under both.
 
\paragraph{Methodology.}
Prior agentic benchmarks report confidence intervals over multiple runs via a normal ($z$) approximation, i.e., estimate $\pm\,1.96\cdot\mathrm{SE}$, e.g., Terminal-Bench~\citep{merrill2026terminalbench} and DeepSWE~\citep{huang2026deepswe}.
Since our number of seeds is small ($n{=}8$), we instead use the Student's $t$-distribution,
\begin{equation}
\bar{x} \;\pm\; t_{0.975,\,\mathrm{df}=7}\cdot \frac{s}{\sqrt{n}},
\end{equation}
where $\bar{x}$ and $s$ are the mean and standard deviation of pass@1 across seeds, yielding more conservative intervals than the normal approximation.

\paragraph{Results.}
As shown in Table~\ref{tab:ci}, the intervals are narrow (widths of roughly $0.02$--$0.05$ in pass@1), indicating that the results are highly stable across seeds despite the limited number of examples.
Crucially, the confidence intervals of our fine-tuned models do not overlap with those of their base counterparts: $[0.2762, 0.3100]$ vs.\ $[0.1400, 0.1807]$ for the 4B pair, and $[0.4040, 0.4235]$ vs.\ $[0.3277, 0.3775]$ for the 9B pair.
This confirms that the improvements reported in \S\ref{sec:experiments} are statistically robust rather than an artifact of the small benchmark size.

\begin{table}[t]
\centering
\small
\setlength{\tabcolsep}{3pt}
\begin{tabular}{@{}p{0.30\columnwidth}p{0.50\columnwidth}r@{}}
\toprule
\textbf{Domain} & \textbf{Platforms} & \textbf{\#Tasks} \\
\midrule
Traffic & Korea Expressway Corporation, Seoul Open Data Plaza & 22 \\
Finance & DART, KRX, Korea Deposit Insurance Corporation & 29 \\
Education & Educational Data Platform, HRD Korea & 30 \\
Law & Korean Law Info Center & 22 \\
Politics & Open Assembly & 25 \\
District Admin. & Gyeonggi Data Dream & 17 \\
\midrule
\textbf{Total} & & \textbf{145} \\
\bottomrule
\end{tabular}
\caption{Selected platforms grouped by domain.}
\label{tab:kopa_platform}
\end{table}

\begin{table*}[t]
\centering
\small
\begin{tabular}{@{}lrrrrrr@{}}
\toprule
\textbf{Domain} & \textbf{\#Tasks} & \textbf{Steps} & \textbf{Tool calls} & \textbf{Headroom} & \textbf{Avail. tools} & \textbf{Parallel \%} \\
\midrule
Traffic & 22 & 3.64 & 5.14 & 2.21 & 7.5 & 55\% \\
Finance & 29 & 3.69 & 5.65 & 2.24 & 6.6 & 83\% \\
Education & 30 & 2.93 & 4.30 & 2.79 & 7.7 & 57\% \\
Law & 22 & 3.00 & 3.55 & 3.63 & 6.7 & 27\% \\
Politics & 25 & 3.72 & 5.48 & 2.10 & 8.7 & 52\% \\
District Administration & 17 & 2.76 & 5.41 & 1.55 & 8.0 & 82\% \\
\midrule
\textbf{Total / mean} & \textbf{145} & \textbf{3.32} & \textbf{4.92} & \textbf{2.45} & \textbf{7.5} & \textbf{59\%} \\
\bottomrule
\end{tabular}
\caption{Per-domain benchmark statistics. Means are weighted by the number of tasks per domain.}
\label{tab:stats}
\end{table*}

\section{Benchmark Statistics}
\label{app:stats}
The benchmark spans 10 platforms across six domains, listed in
Table~\ref{tab:kopa_platform}.
Per-domain statistics characterizing the
structure and difficulty of the 145 evaluation tasks are reported in
Table~\ref{tab:stats}. All quantities are computed directly from the golden action annotations and task metadata.

\paragraph{Metric definitions.}
\begin{itemize}[noitemsep,topsep=2pt]
\item \textbf{Steps} --- the number of sequential reasoning steps required to solve the task. Two golden actions sharing a step index are issued in parallel and count as a single step.
\item \textbf{Tool calls} --- the total number of golden actions in the task. Because parallel actions share a step, the mean tool-call count (4.92) exceeds the mean step count (3.32).
\item \textbf{Headroom} --- the efficiency headroom, defined as $\texttt{max\_allowed\_calls} / N_{\text{opt}}$, where $N_{\text{opt}}$ is the number of golden actions. It quantifies how tight the efficiency constraint is for each task: a value of 1.0 reduces ACTION efficiency to zero after the first redundant call, whereas larger values provide more tolerance before the \texttt{max\_allowed\_calls} ceiling is reached. 
\item \textbf{Avail. tools} --- the mean number of tools exposed to the agent (\texttt{available\_tools}) per task. A larger pool increases the tool selection difficulty.
\item \textbf{Parallel \%} --- the fraction of tasks containing at least one set of parallel calls, i.e., two or more golden actions sharing the same step index. 

\end{itemize}

\section{Execution-Grounded Dynamic Graph: Details and Analysis}
\label{app:app-egdg}
This appendix expands on Phase~A (\S\ref{sec:edge-phasea}). We describe the posterior update scheme (\S\ref{app:egdg_update}), the pruning procedure (\S\ref{app:egdg_hyperparams}), and analyses of graph convergence (\S\ref{app:convergence},\S\ref{app:calibration}). All hyperparameters are listed in Table~\ref{tab:hyperparams}.

\subsection{Update Scheme}
\label{app:egdg_update}

This appendix details the outcome taxonomy and the posterior-update rule
summarized in \S\ref{sec:edge-phasea}, together with the stabilizing mechanisms.
Each edge execution is classified into one of the outcomes in
Table~\ref{tab:taxonomy}, grouped into \emph{structural} faults, which
implicate the binding, and \emph{environmental} faults, which reflect
infrastructure noise independent of the binding. Successes increment the
edge's Beta $\alpha_e$; failures increment $\beta_e$, with structural
faults penalized far more heavily than environmental ones.
The asymmetry between structural and environmental penalties reflects the
public-API environment: rate limits, timeouts, and server errors strike
edges at random, regardless of whether the binding is correct, so they
should only weakly lower an edge's posterior.

\begin{table}[]
\centering
\small
\setlength{\tabcolsep}{4pt}
\begin{tabular}{@{}llc@{}}
\toprule
\textbf{Outcome} & \textbf{Group} & \textbf{Update} \\
\midrule
Semantic success         & ---           & $\alpha_e \mathrel{+}= 1.0$ \\
Generic-only success     & ---           & $\alpha_e \mathrel{+}= 0.3$ \\
\midrule
Schema/type mismatch     & structural    & $\beta_e \mathrel{+}= 1.5$ \\
Missing-field error      & structural    & $\beta_e \mathrel{+}= 1.0$ \\
\midrule
Rate limit               & environmental & $\beta_e \mathrel{+}= 0.1$ \\
Timeout                  & environmental & $\beta_e \mathrel{+}= 0.3$ \\
Server error      & environmental & $\beta_e \mathrel{+}= 0.3$ \\
Authorization error               & environmental & $\beta_e \mathrel{+}= 0.5$ \\
Endpoint unavailable     & environmental & $\beta_e \mathrel{+}= 0.8$ \\
\midrule
Uncategorized failure    & ---           & $\beta_e \mathrel{+}= 0.5$ \\
\bottomrule
\end{tabular}
\caption{Outcome taxonomy and the corresponding Beta-posterior update.
Successes increment $\alpha_e$, more for a \emph{semantic} success
($1.0$) than for a \emph{generic-only} success ($0.3$); structural
failures increment $\beta_e$ by $1.0$--$1.5$, environmental failures by
only $0.1$--$0.8$. A success is \emph{semantic} if at least one
non-generic input parameter (i.e., not an API key, pagination, or format
field) carried a value, and \emph{generic-only} otherwise.}
\label{tab:taxonomy}
\end{table}

\subsection{Pruning Details}
\label{app:egdg_hyperparams}

\begin{figure}[t]
    \centering
    \includegraphics[width=\columnwidth]{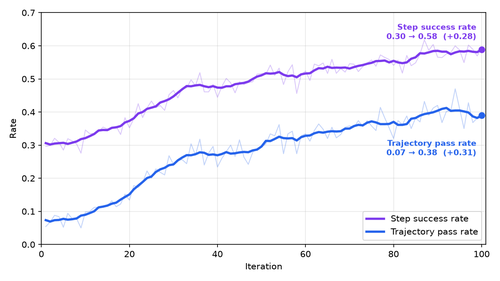}
    \caption{\textbf{Execution success rises as unreliable edges are pruned.}
    Trajectory pass rate and step success rate of sampled paths over the $100$
    Phase~A iterations; faint lines are per-iteration values and bold lines a
    moving average. As the loop prunes structurally failing edges and samples
    toward reliable ones, both rise steadily, indicating that refinement
    concentrates the graph on executable dependencies.}
    \label{fig:convergence-curve}
\end{figure}

\paragraph{Pruning rule.}
An edge $e$ with posterior $\theta_e \sim \mathrm{Beta}(\alpha_e, \beta_e)$
is pruned at iteration $t$ if and only if it has accumulated at least
$n_{\min}$ trials and its posterior mass above the viability threshold
$\tau$ has fallen below the confidence level $\varepsilon$:
\begin{equation}
\label{eq:prune}
\mathrm{prune}(e) \iff
n_e \ge n_{\min}
\;\wedge\;
\Pr\!\left[\theta_e > \tau\right] < \varepsilon,
\end{equation}
where $\Pr[\theta_e > \tau] = 1 - I_{\tau}(\alpha_e, \beta_e)$ is the
complement of the regularized incomplete beta function, and $n_e$ is the
edge's trial count. 

\paragraph{Settings.}
The Beta prior of each admitted skeleton edge is set from the LLM-judged
feasibility score $s_e \in [0,1]$ of its binding, as
$\mathrm{Beta}\!\left(1 + \kappa s_e,\; 1 + \kappa (1 - s_e)\right)$, so
that $\kappa$ controls how strongly the prior commits to the initial
feasibility estimate. An edge enters the skeleton only if
$s_e \ge s_{\min}$. 
During each iteration the sampler draws paths by
Thompson sampling, taking an $\varepsilon_{\mathrm{exp}}$-greedy random edge
with probability $\varepsilon_{\mathrm{exp}}$; edges are then updated
(Table~\ref{tab:taxonomy}) and pruned by Eq.~\eqref{eq:prune}.

\subsection{Graph Convergence Analysis}
\label{app:convergence}
\label{app:graph_analysis}

\paragraph{Execution success rises as the loop prunes unreliable edges.}
The comparisons above contrast edge sets at convergence;
Figure~\ref{fig:convergence-curve} shows the same effect emerging during the
run. As iterations proceed, structurally failing edges accumulate posterior
mass below the pruning threshold and are removed, while Thompson sampling
increasingly routes execution toward edges with demonstrated success. The
trajectory pass rate and step success rate of sampled paths rise steadily as a
result, by $+31$pp and $+28$pp over the $100$ iterations, consistent with the
cross-sectional gap above between the retained graph and the pruned set. The
loop thus progressively concentrates the graph on executable dependencies
rather than reshaping it arbitrarily.

\subsection{Prior Calibration: LLM Judgment versus Execution Evidence}
\label{app:calibration}

\begin{figure}[t]
    \centering
    \includegraphics[width=\columnwidth]{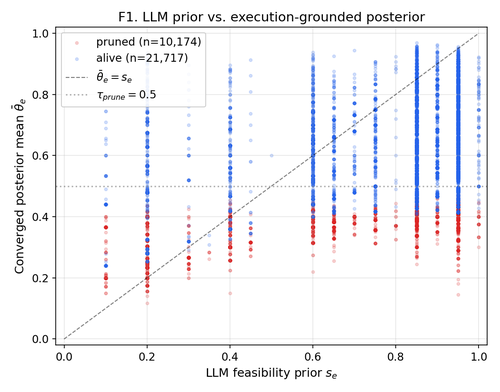}
    \caption{\textbf{Pruning is governed by execution, not by the LLM prior.}
    Each point is an admitted edge, plotting its feasibility prior $s_e$ against
    the converged posterior mean $\bar{\theta}_e$. Retained and pruned edges
    separate along the posterior axis near $\tau = 0.5$ rather than along the
    prior axis, so an edge can carry a high prior yet be pruned once execution
    contradicts it.}
    \label{fig:calibration}
\end{figure}

Prior graph-based methods commit to edge reliability from the LLM's judgment alone. Our loop instead grounds it in execution, and this section shows that both the posteriors and the pruning decisions follow observed outcomes, pruning even highly rated edges when execution contradicts the LLM.
The skeleton anchors each edge's Beta prior to its LLM feasibility score $s_e$ (\S\ref{sec:edge-phasea}), and the loop revises this into a posterior with mean $\bar{\theta}_e$. The revision is worth its cost only if $\bar{\theta}_e$ departs from $s_e$, since a posterior that merely tracked the score would make the loop redundant. Figure~\ref{fig:calibration} plots $\bar{\theta}_e$ against $s_e$ for every admitted edge probed at least once, with the diagonal marking $\bar{\theta}_e = s_e$ and the pruning threshold $\tau$.

\textbf{Execution evidence overrides the prior.} Edges spread broadly along the posterior axis rather than concentrating on the diagonal, so $\bar{\theta}_e$ is driven by observed outcomes rather than $s_e$. The deliberately weak prior $\kappa = 2.0$ lets a few executions move an edge well away from its initial score, and the correction runs both ways, as high-prior edges ($s_e \ge 0.6$) often fall below the diagonal while some low-prior edges ($s_e \le 0.3$) rise above it.

\textbf{Pruning follows the posterior, not the prior.} Because the decision reads $\bar{\theta}_e$ rather than $s_e$, retained and pruned edges interleave across the full range of $s_e$. The boundary runs horizontally near $\tau$, so the two sets separate along the posterior axis, the signature of an execution-grounded decision. The loop thus removes precisely the edges the LLM endorsed but the live APIs reject, which a schema-only pipeline would retain. The cold-start guard ($n_{\min}=2$, Eq.~\ref{eq:prune}) keeps this safe, as none of the $10{,}174$ pruned edges had zero trials.

\begin{table*}[t]
\centering
\small
\begin{tabular}{@{}p{1.6cm} p{2.6cm} p{\dimexpr\textwidth-4.2cm-4\tabcolsep\relax}@{}}
\toprule
\textbf{Stage} & \textbf{Filter Class} & \textbf{Representative Anomaly Example (User Query / Trajectory Flaw)} \\
\midrule
\textbf{Stage 1} & Real-time source & Trajectory components invoke APIs that fetch live highway gating status without a fixed reference timestamp, causing the ground-truth targets in the training dataset to dynamically change per execution (violating data determinism). \vspace{0.15cm} \par
\emph{Example:} ``실시간 영업소 진입조절 현황에 뜬 각 노선별로 휴게시설 경로 정보가 몇 건씩 있는지 확인해 줄 수 있어?'' \vspace{0.05cm} \par
(\emph{Gloss:} ``Can you check how many rest area route information items are available for each route listed in the real-time tollbooth entry control status?'') \\
\midrule
\textbf{Stage 1} & Linguistic anomaly & Query contaminated with foreign tokens or mixed scripts during synthesis, violating standard format constraints. \vspace{0.15cm} \par
\emph{Example:} ``전월 대비总人口 증감 수치를 알려줘." or "먼저 확인한 뒤,そこに 포함된 각 영업소별로''. \vspace{0.05cm} \par
(\emph{Gloss:} ``Tell me the change in 总人口 compared to the previous month." / "After checking first, そこに for each tollbooth included there.'') \\
\midrule
\textbf{Stage 2} & Answer leakage & The query phrasing explicitly discloses the target intermediate parameter, rendering the associated tool call redundant. \vspace{0.15cm} \par
\emph{Example:} ``2023년 김포시에 있는 야구 가능한 공공체육시설 주소에서 시군명을 뽑아서, 그 시군별로 경기도 내 골프장이 몇 건씩 있는지 알려줄래?'' \vspace{0.05cm} \par
(\emph{Gloss:} ``Extract the municipality name from the address of baseball-available public sports facilities in Gimpo City in 2023, and tell me how many golf courses there are in Gyeonggi-do for that municipality?'') \\
\bottomrule
\end{tabular}
\caption{Taxonomy and representative examples of filtered data anomalies, categorized by filtering stage, filter class, and associated query/trajectory flaws.}
\label{tab:filtered-examples}
\end{table*}

\begin{table}[h]
\centering
\small
\setlength{\tabcolsep}{4pt}
\begin{tabularx}{\columnwidth}{@{}l >{\raggedright\arraybackslash}X c@{}}
\toprule
\textbf{Symbol} & \textbf{Description} & \textbf{Value} \\
\midrule
\multicolumn{3}{@{}l}{\emph{Phase~A: skeleton graph construction}}\\
$K$               & Intra-domain candidate neighbors / source tool & $15$ \\
$K_{\mathrm{x}}$  & Cross-domain candidate neighbors / source tool & $10$ \\
$s_{\min}$        & Min.\ feasibility score to admit an edge  & $0.3$ \\
$\kappa$          & Prior strength                            & $2.0$ \\
                  & Skeleton-construction model              & Qwen3.5-122B \\
\midrule
\multicolumn{3}{@{}l}{\emph{Phase~A: execution-grounded graph update}}\\
$T$               & Graph-update iterations                  & $100$ \\
$M$               & Paths sampled per iteration              & $500$ \\
$\varepsilon_{\mathrm{exp}}$ & $\varepsilon$-greedy exploration rate & $0.1$ \\
$\tau_{\mathrm{prune}}$      & Viability threshold (Eq.~\ref{eq:prune}) & $0.5$ \\
$\varepsilon_{\mathrm{prune}}$ & Pruning confidence level (Eq.~\ref{eq:prune}) & $0.3$ \\
$n_{\min}$        & Min.\ trials before pruning              & $2$ \\
\midrule
\multicolumn{3}{@{}l}{\emph{Phase~B: trajectory synthesis}}\\
$\phi$            & Fan-out budget (\S\ref{sec:method:trace}) & $5$ \\
\bottomrule
\end{tabularx}
\caption{\textsc{EDGE} hyperparameters and the values used in all
reported runs.}
\label{tab:hyperparams}
\end{table}

\section{Data Filtering Details}
\label{app:filter}

To optimize efficiency, we filter the synthesized corpus with a three-stage cascading pipeline with increasing computational costs. Table~\ref{tab:filtered-examples} presents representative examples of execution flaws, textual anomalies, and logical leaks identified and pruned during synthesis.

\textbf{Stage 1 (Rule-based filters).} Deterministic rules remove four types: \emph{non-reproducible} instances (e.g., real-time queries without a fixed timestamp), \emph{erroneous} data whose ground truth is an error signal, \emph{redundant} near-duplicates, and \emph{malformed} instances violating format or schema constraints. 

\textbf{Stage 2 (LLM-based static filters).} An LLM judge then removes ill-posed instances: \emph{non-gradable} queries whose correctness cannot be verified automatically, \emph{answer-leaking} queries that disclose the target answer, and \emph{decorative chains} whose intermediate calls do not constrain the final answer. The latter two require agreement between two independent LLM judges. 

\textbf{Stage 3 (Independent re-solving).} Neither stage detects errors in the ground truth itself. Whereas existing pipelines treat it as a fixed reference~\citep{liu2024apigen}, we re-solve every task independently; a re-solution judged correct actively replaces the original stored answer. Throughout this pipeline, only open-source model outputs become training labels and proprietary models serve solely for verification.

\section{Training Data Analysis}

\paragraph{Training data statistics.}
\label{app:corpus}
Each task is typed by the Phase~B
taxonomy of \S\ref{sec:method:trace}, and Table~\ref{tab:type-distribution}
reports the resulting training data distribution. The key effect of the cardinality-based
junction typing is visible among the single-junction sequential
trajectories. The high-cardinality fields of Korean public APIs
(\S\ref{sec:method:trace}) make one-to-many junctions the norm, so that
\textsc{Fan} ($2{\le}n_i{\le}\phi$) and \textsc{Drv} ($n_i{>}\phi$) together
account for $64.9\%$ of these trajectories ($398$ of $613$), far outnumbering
the one-to-one \textsc{Seq} junctions ($n_i{=}1$, $35.1\%$) to which a naive
one-to-one template would be confined. The remaining tasks are template-based
non-sequential types (\textsc{Sem}, \textsc{Cmp}, \textsc{Cond}, $28.9\%$ of
the corpus) and Mixed tasks ($36.7\%$) that combine more than one of
these patterns within a single trajectory.

Beyond type composition, the corpus is structurally demanding.
Table~\ref{tab:fc-stats} reports its call structure over the $1{,}781$ tasks.
A task issues $4.13$ tool calls on average and up to $36$, and $57.6\%$ of
tasks require at least four sequential hops, so a model must sustain long
chains of dependent calls rather than answer in a single step. The structural depth, together with the one-to-many chaining
detailed in Table~\ref{tab:chaining_statistics}, is what the model must learn to
handle.

\paragraph{Comparison with existing tool-use datasets.}
\label{app:comparison}
\begin{table}[t]
\centering
\small
\setlength{\tabcolsep}{2pt}
\begin{tabular*}{\columnwidth}{@{}l@{\extracolsep{\fill}}r@{}}
\toprule
\textbf{Metric} & \textbf{Value} \\
\midrule
Calls per task (mean / med. / min / max) & 4.13 / 4 / 2 / 36 \\
Steps: $\leq$3 / 4--5 / $\geq$6 & 42.3 / 50.9 / 6.7\,\% \\
Calls: sequential / parallel & 63.8 / 36.2\,\% \\
\bottomrule
\end{tabular*}
\caption{Training-data statistics over the $1{,}781$ synthesized tasks:
the distribution of calls per task, the share of tasks by step count, and
the split of calls into sequential and parallel.}
\label{tab:fc-stats}
\end{table}

What distinguishes our corpus from prior tool-use datasets is the prevalence
of one-to-many chaining, a direct consequence of the high-cardinality fields
of Korean public APIs. Table~\ref{tab:chaining_statistics} compares our corpus
against representative tool-use datasets: $74.1\%$ of calls in our corpus
return at least two records and $81.2\%$ of chained calls are one-to-many,
with a median chained cardinality of $27$ (up to $224{,}958$), whereas the
prior datasets stay below $5\%$ one-to-many chaining with single-digit
cardinality. This is precisely the regime that the cardinality-based junction
typing of \S\ref{sec:method:trace} is designed to handle, and that a
one-to-one synthesis template cannot reach.

\begin{table*}[t]
\centering
\small
\begin{tabular}{@{}llrrrr@{}}
\toprule
\multirow{2}{*}{\textbf{Dataset}} &
\multirow{2}{*}{\textbf{Environment}} &
\multirow{2}{*}{\textbf{one-to-many chaining (\%)}} &
\multicolumn{3}{c}{\textbf{chained cardinality}} \\
\cmidrule(l){4-6}
& & & \textbf{median} & \textbf{p90} & \textbf{max} \\
\midrule
\textbf{ToolBench-v1} & RapidAPI  &  1.9 &  1 &  16 & 127 \\
\textbf{APIGen-MT}    & $\tau$-bench & 10.0 &  1 &   3 &  16 \\
\textbf{Nemotron} ~\citep{nvidia2025nemotronsftagenticv2}     & $\tau^2$-bench &  1.6 &  1 &   1 &  15 \\
\textbf{ToolACE}      & synthetic &  4.5 &  1 &   5 &  10 \\
\textbf{Ours}         & \textbf{\textsc{KOPA-Bench}} & \textbf{81.2} & \textbf{27} & \textbf{157} & \textbf{224{,}958} \\
\bottomrule
\end{tabular}
\caption{Comparison of chaining-related statistics across datasets. \emph{one-to-many chaining}
is the share of multi-step trajectories that contain at least one junction consuming a
multi-record tool output. \emph{chained cardinality} reports the distribution of records consumed
at such junctions.}
\label{tab:chaining_statistics}
\end{table*}

\begin{table}[t]
  \centering
  \small
  \setlength{\tabcolsep}{6pt}
  \begin{tabular}{lrr}
    \toprule
    Type & Count & \% \\
    \midrule
    \textsc{Mixed} & 654  & 36.7 \\
    \textsc{Drv}   & 350  & 19.7 \\
    \textsc{Sem}  & 303  & 17.0 \\
    \textsc{Seq}   & 215  & 12.1 \\
    \textsc{Cmp}   & 161  & 9.0  \\
    \textsc{Cond}  & 50   & 2.8  \\
    \textsc{Fan}   & 48   & 2.7  \\
    \midrule
    Total & 1781 & 100.0 \\
    \bottomrule
  \end{tabular}
  \caption{Distribution of task types in training set}
  \label{tab:type-distribution}
\end{table}

\section{Data Contamination Audit}
\label{app:contamination}

Because \textsc{KOPA-Bench} and the \textsc{EDGE} training corpus are both built
on the same live public APIs, they necessarily draw on a shared pool of tools.
We therefore audit how much the 145 evaluation tasks overlap with the 1{,}781
training tasks, separating genuine leakage from the shared pool that
a function-calling benchmark and its training corpus are expected to share.
We define \emph{tool-universe overlap} as the fraction of distinct evaluation
gold functions that also appear as a gold function anywhere in training. This
measures the shared API pool, which is intended by design and is distinct
from query or task leakage. We report the audit at five levels, summarized in
Table~\ref{tab:datacontamination}. 
No evaluation query appears verbatim in training (0 of 145), so there is no
direct query leakage. The tool-universe overlap is 62.2\% (135 of 217 gold
functions), which reflects the shared API pool rather than contamination.

\begin{table}[t]
\centering
\small
\begin{tabular}{@{}lr@{}}
\toprule
\textbf{Level} & \textbf{Value} \\
\midrule
Verbatim query leakage                 & 0.0\% (0/145) \\
Tool-universe overlap (functions)      & 62.2\% (135/217) \\
Class-level held-out platforms           & 3/12 \\
Tasks using only unseen functions      & 12.4\% (18/145) \\
Mean per-task function coverage        & 62.9\% \\
\bottomrule
\end{tabular}
\caption{Train--eval contamination audit between \textsc{KOPA-Bench} (145 tasks)
and the \textsc{EDGE} training corpus (1{,}781 tasks).}
\label{tab:datacontamination}
\end{table}

Verbatim query leakage at 0\% is the strongest available evidence that the benchmark
is not contaminated by the training corpus.

\subsection{Held-Out Platform Evaluation}
\label{app:heldout}

Tool-universe overlap alone does not show whether the reported gains depend on
the shared pool. We evaluate on the subset of \textsc{KOPA-Bench}
grounded in platforms that never enter synthesis: Seoul Open Data Plaza in the
Traffic domain, and DART and KRX in the Finance domain. These three platforms of
Table~\ref{tab:kopa_platform} appear only in \textsc{KOPA-Bench}, and none of
their API tools appears in any training trajectory, so the held-out unit is the
platform rather than the individual function. In total, 31 benchmark tasks are
grounded in them, 10 on Seoul Open Data Plaza and 21 on the two finance
platforms.

The only tool these tasks share with training is
\texttt{evaluate\_expression}, the general-purpose evaluator of
Appendix~\ref{app:task-example}, which takes an arithmetic or symbolic
expression as a string and returns its value. Many tasks close with an
arithmetic step and so call it, but it queries no external service and carries
no platform-specific API knowledge. Once it is set aside as
a general-purpose utility, every public-API tool involved in the 31 tasks is
absent from training.

Table~\ref{tab:heldout} reports pass@4 on this subset, as defined in
Table~\ref{tab:main}. Fine-tuning improves performance on the held-out platforms
rather than degrading it, from $0.2903$ to $0.5161$ ($9$ to $16$ of the $31$
tasks), a gain of $+22.6$pp that exceeds the $+15.9$pp the same model gains over the full benchmark ($0.3103 \rightarrow 0.4690$). The improvement is therefore not
confined to the platforms seen during synthesis, as it would be if the model had
acquired tool-chain templates or platform-specific call procedures from
training.

\begin{table}[t]
\centering
\small
\setlength{\tabcolsep}{4pt}
\begin{tabular}{@{}llrrr@{}}
\toprule
\textbf{Domain} & \textbf{Held-out platform} & $n$ & \textbf{Base} & \textbf{Ours} \\
\midrule
Traffic & Seoul Open Data Plaza & 10 & 0.2000 & 0.6000 \\
Finance & DART•KRX & 21 & 0.3333 & 0.4762 \\
\midrule
\multicolumn{2}{@{}l}{\textbf{Total}} & \textbf{31} & \textbf{0.2903} & \textbf{0.5161} \\
\bottomrule
\end{tabular}
\caption{Performance on the 31 \textsc{KOPA-Bench} tasks grounded in platforms
held out from synthesis, for Qwen3.5-4B before and after training on the
\textsc{EDGE} corpus. $n$ is the number of tasks per platform; values are
pass@4 as defined in Table~\ref{tab:main}.}
\label{tab:heldout}
\end{table}

\subsection{Dependency Edge Overlap}
\label{app:edgeoverlap}

The audits above compare tool inventories and whole tasks. 
A model could still benefit from having seen the same dependency structure during training 
even when the individual tools differ, so we audit overlap at the level of edges between calls, at two resolutions. A \emph{function-pair edge} $(u, v)$ records that an output field of tool $u$ supplies an input parameter of tool $v$, which is the dependency relation that \textsc{EDGE} synthesizes and that the model must recover at evaluation time. An \emph{adjacency edge} links two consecutive calls in a trajectory regardless of whether a dependency is present, serving as a loose upper bound on structural overlap.

We count every edge occurrence across all 145 evaluation tasks without deduplication, so a structure repeated across tasks contributes proportionally to how often the model encounters it. The Jaccard similarity is computed over sets of unique edges, 
$|\mathcal{E}_{\text{eval}} \cap \mathcal{E}_{\text{train}}| /
 |\mathcal{E}_{\text{eval}} \cup \mathcal{E}_{\text{train}}|$, and \emph{Unseen} denotes the fraction of edge occurrences in evaluation whose edge never appears in training. Edges that merely acquire an API key are excluded: in the open-API setting of \textsc{KOPA-Bench}, key acquisition precedes every call and is thus a fixed property of the environment rather than a task-specific structure. At the dependency resolution, this exclusion is automatic, since an API key is a generic parameter filled through the \textsc{Default} branch of Eq.~(\ref{eq:sourceplan}) rather than \textsc{Upstream}
(\S\ref{sec:edge-phasea}), and therefore never forms a dependency edge.

Table~\ref{tab:edgeoverlap} shows that the dependency edges do not overlap at all.
The Jaccard similarity is 0.0, and all 178 evaluation dependency-edge occurrences are unseen in training, despite the $62.2\%$ tool-universe overlap. That is, the benchmark and the training corpus draw on a shared pool of tools but connect them in different ways. Even under the loose adjacency bound, $95.6\%$ of occurrences remain unseen, so this conclusion does not depend on how strictly an edge is defined.

\begin{table}[t]
\centering
\small
\setlength{\tabcolsep}{4pt}
\begin{tabular}{@{}lrrrr@{}}
\toprule
\textbf{Resolution} & \textbf{Eval} & \textbf{Train} & \textbf{Jaccard} & \textbf{Unseen} \\
\midrule
Function-pair & 178 & 687   & 0.000 & $100\%$ \\
Adjacency     & 592 & 2{,}911 & 0.004 & $95.6\%$ \\
\bottomrule
\end{tabular}
\caption{Overlap of edges between calls in \textsc{KOPA-Bench} and the
\textsc{EDGE} training corpus. Eval and Train are edge occurrence counts without
deduplication, while Jaccard is computed over the corresponding sets of unique
edges. Function-pair is the dependency relation; adjacency links any two
consecutive calls and serves as an upper bound. API key acquisition edges are
excluded.}
\label{tab:edgeoverlap}
\end{table}

Together with \S\ref{app:heldout}, the audits point in the same direction. The shared tool pool is intended by design, yet the platforms behind 31 evaluation tasks never enter synthesis, and the dependency structure the model must recover at evaluation time is entirely new.

\section{Training Details}
\label{app:training}
We fine-tune Qwen3.5-4B (dense, 28-layer, bfloat16) with Group
Relative Policy Optimization (GRPO) on the multi-step tool-calling corpus
of \S\ref{sec:datagen}, using verl with a Megatron-LM policy
backend and a vLLM rollout backend on a single $8\times$NVIDIA H100
80\,GB node.
The GRPO objective uses a train batch of $8$, aggregates the loss by
token-mean, and applies a $k_3$ KL penalty of $0.03$ to the loss with no
KL term in the reward. We use a context length of $65$K tokens. For the
multi-turn agent, we cap each episode at $15$ assistant turns and $16$K
tool-response tokens, with episode and tool timeouts of $600$\,s and
$60$\,s.

\section{Prompts}
\label{app:method-prompts}
This section provides the prompt templates used by the EDGE synthesis
pipeline (\S\ref{sec:datagen}): dependency extraction and argument
generation in Phase~A (\S\ref{app:prompt-phasea}), query generation in
Phase~B (\S\ref{app:prompt-phaseb}), and the LLM-based static filter
(\S\ref{app:prompt-filter}).

\subsection{Phase A: Graph Construction}
\label{app:prompt-phasea}
Phase~A invokes two prompts. Dependency Extraction
(Table~\ref{tab:prompt-binding}) extracts dependencies when building the
skeleton graph, returning the feasibility score $s_e$, the binding set
$B_e$, and the default candidates $D_e$ of Eq.~\eqref{eq:sourceplan} in a single
structured response; feasibility scoring is thus not a separate call but
the \texttt{feasibility\_score} field of this prompt.
Source-Argument Generation (Table~\ref{tab:prompt-srcarg})
generates valid argument values for the source tool when an edge is probed
against the live API, with API key and pagination parameters excluded and
injected separately. Both prompts are run with Qwen3.5-122B.

\subsection{Phase B: Trajectory and Query Synthesis}
\label{app:prompt-phaseb}

Phase~B generates a Korean query for each completed trajectory whose
answer requires the full tool-call chain. All Korean prompts are shown in
English translation. All calls share a system prompt
(Table~\ref{tab:prompt-phaseb-system}) that enforces a conversational
tone, prohibits procedural and API references, restricts value citations
to seed call arguments or process-step filter values, and mandates a
five-item self-check before output.
The user prompt varies by trajectory type (\S\ref{sec:method:trace}).
\textit{Sequential} trajectories (\textsc{Seq}, \textsc{Fan}, \textsc{Drv}
and compositions) use one template (Table~\ref{tab:prompt-phaseb-chain})
with a pattern-specific assembly rule. \textit{Non-sequential}
trajectories (\textsc{Sem}, \textsc{Cmp}, \textsc{Cond}) use two stages, a per-step sub-query prompt (Table~\ref{tab:prompt-phaseb-subquery}) and a
composition prompt (Table~\ref{tab:prompt-phaseb-compose}) that keeps
every step logically necessary. A separate answer prompt
(Table~\ref{tab:prompt-phaseb-answer}) handles these, with per-target
enumeration for \textsc{Sem} and \textsc{Cmp} and two-sentence branch
selection for \textsc{Cond}. Phase~B synthesis uses both Qwen3.5-122B and Qwen3.5-397B-A17B.

\subsection{Data Filtering}
\label{app:prompt-filter}
The static filtering pipeline deploys dedicated LLM classifiers to detect and eliminate three categories of ill-posed data: (a)~\emph{non-gradable} queries, (b)~\emph{answer-leaking} queries, and (c)~\emph{decorative chains}. We present the complete system prompt utilized for detecting \emph{answer-leaking} queries in Table~\ref{tab:prompt-data-filtering} as the representative example of our filtering rubrics. The validation stage
uses Qwen3.5-397B-A17B and GPT-5~\citep{singh2026openaigpt5card} as the referee.

\subsection{Evaluation Prompts}
\label{app:eval-prompts}
This section provides the prompts used during evaluation
(\S\ref{sec:experiments}). The agent under test receives the system prompt
of Table~\ref{tab:prompt-agent}. When deterministic answer extraction
fails, RESPONSE grading falls back to an LLM judge, using a shared
prompt for the \textit{string} and \textit{judge} response types
(Table~\ref{tab:prompt-judge-string}) and a separate prompt with
numeric-equivalence rules for the \textit{number} type
(Table~\ref{tab:prompt-judge-number}).

\onecolumn
\begin{tcolorbox}[
  breakable,
  colback=white,
  colframe=black,
  colbacktitle=gray!20,
  coltitle=black,
  title={\textit{\textbf{Prompt for Dependency Extraction}}},
  fonttitle=\normalsize,
  boxrule=0.4pt,
  left=3pt, right=3pt, top=2pt, bottom=2pt,
  halign title=center
]
\begin{lstlisting}[basicstyle=\ttfamily\scriptsize, breaklines=true, columns=fullflexible, keepspaces=true, language={}, showstringspaces=false, frame=none]
You are given two API functions: a **Source** function and a **Target** function.
Your task is to:
1. Evaluate the structural feasibility of using Source's output as input to Target.
2. Extract explicit **binding sets**: which output fields of Source map to which input parameters of Target.

Source API Function:
{source_api_str}

Target API Function:
{target_api_str}

**Instructions**:
- For each Target input parameter (EXCLUDING common request params like KEY, TYPE, START_INDEX, END_INDEX, dataFormat, numOfRows, pageNo, pIndex, pSize, service, resultType, OC, api_key, service_key, serviceKey), determine if any Source output field can provide its value.
- If a transformation is needed (e.g., date format change, type casting), describe it briefly in the "transform" field.
- Assign a confidence score (0.0-1.0) to each binding.
- Also provide an overall feasibility_score (0.0-1.0) using these guidelines:
  - 0.9-1.0: Direct, clear output-to-input dependency
  - 0.7-0.8: Strong dependency, minor transformation needed
  - 0.4-0.6: Moderate dependency, uncertain match
  - 0.2-0.3: Weak, tangential relationship
  - 0.0-0.1: No meaningful dependency

**CROSS-DOMAIN BINDING - CRITICAL**:
Field names often differ across domains even when they represent the SAME concept.
You MUST look at the **description/meaning** of fields, NOT just their names.
If two fields represent the same real-world concept, they SHOULD be bound even if names differ.

Common cross-domain mappings to look for:
- Region/location: SIGUN_NM, REGION_NM, rgn, rgnSeNm, ctpvNm, [region name] -> all mean "region name"
- Region code: SIGUN_CD, REGION_CD, rgnCd -> all mean "region code"
- Year: SUM_YY, fyr, crtrYr, baseyy, YEAR, srchyear -> all mean "year"
- Name/title: BILL_NAME, CNTNTS_TITLE, schlNm, opnId -> entity identifiers
- Date: USE_YMD, BGN_DE, END_DE -> date values

Example cross-domain bindings:
  education.output["ctpvNm"] -> hrdk.input["rgn"]  (both = region name, confidence: 0.7)
  national_assembly.output["BILL_NO"] -> law.input["MST"]  (both = bill identifier, confidence: 0.6)
  dataseoul.output["USE_YMD"] -> expressway.input["exDate"]  (both = date, confidence: 0.7)

Do NOT reject a binding just because field names look different.
DO reject a binding only if the semantic meaning is clearly incompatible.

**Directional Constraint**: Evaluate ONLY Source -> Target direction.

**Suggested Defaults for Unbound Parameters**:
For Target input parameters that are NOT bound to any Source output field (excluding KEY/TYPE/pagination),
suggest 3 realistic default value candidates that would make the API call succeed.
This is critical for parameters that would otherwise require user input (e.g., region names, dates, IDs).
Use your knowledge of the API's domain to suggest DIVERSE values that are likely to return non-empty results.
For example, if the parameter is a region name, suggest 3 different regions.

Return as JSON:
{
    "feasibility_score": 0.75,
    "reason": "Brief explanation of the feasibility assessment",
    "bindings": [
        {
            "source_field": "field_from_source_output",
            "target_param": "param_in_target_input",
            "transform": null,
            "confidence": 0.9
        }
    ],
    "suggested_defaults": {
        "unbound_param_1": ["candidate_1", "candidate_2", "candidate_3"],
        "unbound_param_2": ["candidate_1", "candidate_2", "candidate_3"]
    }
}
If no meaningful bindings exist, return an empty bindings list and a low feasibility_score.
\end{lstlisting}
\end{tcolorbox}
\captionof{table}{Prompt for dependency extraction in Phase~A. A single call returns the feasibility score, binding set, and default candidates of Eq.~\eqref{eq:sourceplan}.}
\label{tab:prompt-binding}
\twocolumn

\onecolumn
\begin{tcolorbox}[
  breakable,
  colback=white,
  colframe=black,
  colbacktitle=gray!20,
  coltitle=black,
  title={\textit{\textbf{Prompt for Source-argument generation}}},
  fonttitle=\normalsize,
  boxrule=0.4pt,
  left=3pt, right=3pt, top=2pt, bottom=2pt,
  halign title=center
]
\begin{lstlisting}[basicstyle=\ttfamily\scriptsize, breaklines=true, columns=fullflexible, keepspaces=true, language={}, showstringspaces=false, frame=none]
You are generating valid arguments for a Korean public API tool call.

Tool Name: {tool_name}
Tool Description: {tool_description}
Input Parameters Schema:
{input_schema}

Rules:
- Generate realistic, valid argument values for each parameter.
- DO NOT generate values for API key parameters (KEY, api_key, service_key, serviceKey) - these will be injected automatically.
- DO NOT generate values for pagination/format parameters (numOfRows, pageNo, dataFormat, type, start_index, end_index) - these are handled separately.
- For year parameters, use "2023" unless the description suggests otherwise.
- For region/location parameters, use a common Korean region (e.g., Seoul, Busan, Gyeonggi).
- For school-related parameters, leave as empty string "" to get all results.
- For optional parameters with no clear default, use empty string "".
- Return ONLY the parameters that should be included in the API call.

Return as JSON:
{
    "arguments": {"param1": "value1", "param2": "value2"}
}
\end{lstlisting}
\end{tcolorbox}
\captionof{table}{Prompt for source-argument generation in Phase~A, used when probing an edge against the live API.}
\label{tab:prompt-srcarg}
\twocolumn
 
\onecolumn
\begin{tcolorbox}[
  breakable,
  colback=white,
  colframe=black,
  colbacktitle=gray!20,
  coltitle=black,
  title={\textit{\textbf{Query-Generation System Prompt}}},
  fonttitle=\normalsize,
  boxrule=0.4pt,
  left=3pt, right=3pt, top=2pt, bottom=2pt,
  halign title=center
]
\begin{lstlisting}[basicstyle=\ttfamily\scriptsize, breaklines=true, columns=fullflexible, keepspaces=true, language={}, showstringspaces=false, frame=none]
You are a query generator that writes Korean questions and precise answers in the **tone of a real user asking a chatbot/LLM**.

## User perspective
The asker:
- Has domain knowledge (e.g., expressways, rest areas, routes, committees, bills)
- Does NOT know about APIs, functions, data structures, or field names
- Does NOT care how the data is retrieved -- only what they want to know
- Asks in a **natural conversational tone typical of everyday chatbot/LLM usage**, not in a report or paper style

## Tone (conversational) -- applied to the query only
The question must take the form **a real user would send to a chatbot/LLM**.
Formal or report-like tone must be avoided.
**Vary sentence-ending forms** -- do not repeat one form:
- "...How many items?" / "...How many cases?" / "...What is it like?"
- "Please tell me..." / "Tell me..." / "Could you tell me?"
- "I'm curious about..." / "I'd like to know..." / "I wonder..."
**Natural segmentation is encouraged**: do not cram all conditions into one sentence; if it can be split into short sentences, splitting is OK.
**Conversational markers (optional, do not overuse)**: "a little", "just", "by any chance", "by the way", etc.
(Four natural-vs-unnatural example pairs omitted -- contrast between report tone and chatbot tone.)
**Hard rules**:
- No English API field names
- No procedural/mechanism vocabulary ("retrieve", "API", "endpoint", etc.)
- No exposure of meta-processing (tie-breaking, sorting steps, etc.)

## Handling tool descriptions
Input tool descriptions may contain procedural/mechanism vocabulary.
This is implementation detail and must NEVER appear in the question.
- Procedural vocabulary: "retrieve", "in order to retrieve", "search", "verify", "call", "fetch", "obtained from", "shown in", "extracted from"
- Mechanism vocabulary: "OpenAPI", "API", "function", "endpoint", "response"
- Data-processing vocabulary: "the retrieved result", "the returned data", "the relevant list", "result code", "response data"
- English API field names (code-style): SIGUN_NM, BLL_NM, UNIT_CD,
  CONF_ID, etc. -- convert to the Korean description shown in parentheses
- Vague placeholders: "a specific X", "the relevant X", "some X" -- if a concrete value is in the args, it MUST be used

## Handling parenthetical clarifications in tool descriptions
(A) Short formal title / source label -- keep
- "Fund Annual Current Balance (Settlement) (Fiscal Scale)" -- standard Korean statistical naming convention
- Typically <= 10 characters, no commas
(B) Long supplementary clarification -- strip, keep only the core entity
- "... status information (engineers, technicians, etc., applicant qualifications, application procedure, ...)" -> use only "... status information" in the query
- Typically > 10 characters or contains commas
Rule: **If the parenthetical contains even one comma, it is almost always type (B) supplementary -- strip the whole parenthetical**.

## Avoid repeating organizational prefix in tool names
When two tools in the chain share the same source/organization prefix, use the prefix **only once** in the query.

## No exposure of meta-processing or search mechanism
The query is **what the user asks**. System processing details (tie-breaking, sorting, selection steps) must **NEVER appear** in the query.

## Query value citation rule (mandatory)
**Concrete values** appearing in the query (years, codes, region names, thresholds, quarters, etc.) must be one of:
1. A value in the seed step's call args
2. The filter_value of a process step (DRV threshold)
3. All other values -- forbidden in the query (metadata field values from result_rows, guesses based on world knowledge, etc.)
Information that does not qualify must be expressed in generalized form: "for each X", "by region", "matching a certain condition", etc.

## Handling long memo / text values
Long text in seed args (> 30 characters; long combined memo/description/address) must NOT be copied verbatim into the query. Either generalize or extract a short essential portion.

## Naturalize answer fields (answer_fields)
When answer fields are raw English / special expressions, use generalized or unified expressions in the query.

## Self-check before emitting output (mandatory)
After drafting the query, verify the following five items before output:
1. Zero implementation-detail traces: no procedural vocabulary, no API field names, no mechanism vocabulary, no data-processing vocabulary
2. Legal value citation: every concrete value (year / code / region /threshold) exists in the trace's seed args or in a process step's filter_value
3. Parenthetical cleanup: comma-separated list-style parentheticals from tool descriptions were not copied verbatim
4. No prefix repetition: the same source/organization prefix does not appear twice in one query
5. No meta exposure: system rules such as tie-breaking, sorting, or processing steps do not appear in the query
If any of the above is violated, **revise before output**.

## Output format
JSON:
{"query": "natural Korean question from the user's perspective",
 "answer": "concrete answer grounded in execution results",
 "answer_fields": ["field1", "field2"]}
- answer_fields: list of field names in the result records from which the answer values were taken (empty array when only a count is answered).
\end{lstlisting}
\end{tcolorbox}
\captionof{table}{System prompt for the Phase~B query generator, which writes a Korean question in the tone of a real user without exposing any implementation detail.}
\label{tab:prompt-phaseb-system}
\twocolumn

\onecolumn
\begin{tcolorbox}[
  breakable,
  colback=white,
  colframe=black,
  colbacktitle=gray!20,
  coltitle=black,
  title={\textit{\textbf{Chain Query--Answer Assembly}}},
  fonttitle=\normalsize,
  boxrule=0.4pt,
  left=3pt, right=3pt, top=2pt, bottom=2pt,
  halign title=center
]
\begin{lstlisting}[basicstyle=\ttfamily\scriptsize, breaklines=true, columns=fullflexible, keepspaces=true, language={}, showstringspaces=false, frame=none]
Based on the following chain components and execution results, write a query+answer pair.

## Chain composition
### 1. Data source (do NOT cite source vocabulary -- only entities and fields)
- {seed_desc}
{seed_args_section}
### 2. Step connection
{bridge_section}
### 3. Downstream conditions
{downstream_section}
### 4. Question type
{answer_section}

## Execution results
{execution_results}

## Assembly rule
{pattern_assembly_rule}

## Chain-specific additional rules
(In addition to the general rules in the system prompt, extra constraints specific to this chain's composition.)
- **Actual value of the connected field**: in the query express only meaning (e.g., "for each city name"); in the answer include the actual value (e.g., "Yangju-si: 12 cases, Seongnam-si: 5 cases")
- **Bridge value is a chain shortcut**: if the user knew that value they could skip the source step -> NEVER expose it in the query
## Input value naturalization (apply whenever an argument value is mentioned in the query)
- Term/session codes -> natural language: "21" -> "21st", "414" -> "414th session"
- Date codes -> natural language: "2023-04" -> "April 2023", "20230401" -> "April 1, 2023"
- Year -> natural language: "2023" -> "year 2023"
- Quarter -> natural language: "1" -> "Q1", "4" -> "Q4"
## Output (JSON)
{
  "query": "natural Korean question",
  "answer": "concrete answer grounded in execution results",
  "answer_fields": ["field1", "field2"]
}
Answer rules:
- Count question -> exactly reflect the count from execution results (e.g., "Total of 18 cases")
- Field-value question -> extract the actual value from execution results
- No results -> "There is no data matching this condition"
- No placeholders such as "can be verified", "a list is provided", etc.
- **When there are two or more targets, separate each target on its own bullet/numbered line**:
  example: "- Yangju-si: 1504 cases" / "- Seongnam-si: 1504 cases"
\end{lstlisting}
\end{tcolorbox}
\captionof{table}{Prompt that assembles a full query--answer pair from the chain
components and execution results in Phase~B.}
\label{tab:prompt-phaseb-chain}
\twocolumn

\onecolumn
\begin{tcolorbox}[
  breakable,
  colback=white,
  colframe=black,
  colbacktitle=gray!20,
  coltitle=black,
  title={\textit{\textbf{Sub-Query Generation}}},
  fonttitle=\normalsize,
  boxrule=0.4pt,
  left=3pt, right=3pt, top=2pt, bottom=2pt,
  halign title=center
]
\begin{lstlisting}[basicstyle=\ttfamily\scriptsize, breaklines=true, columns=fullflexible, keepspaces=true, language={}, showstringspaces=false, frame=none]
You generate a natural Korean sub-query for a SINGLE tool call step.
## Input
- Tool: {tool_name}
- Description: {tool_description}
- Arguments (MUST be included in the sub-query in natural Korean): {arguments_str}
- Result Summary: {result_summary_str}
{step_context}

## Steps (follow in order)
### Step 1: Identify terminology
Use the tool description as the domain term directly (it is already short).
e.g., "expressway real-time traffic information" -> domain term = "expressway real-time traffic"
### Step 2: Convert arguments to Korean
{arg_handling_rule}
### Step 3: Determine question type
{question_type_rule}
### Step 4: Write the sub-query
Combine the domain term + converted arguments + question type into one natural Korean question.

## FORBIDDEN (any of these -> invalid output)
- API field names as-is: ROUTE_CD, STN_NM, ORG_CD, YEAR, era_co, etc.
- Procedural language: "retrieve", "search", "call", "verified"
- Vague terms: "recent", "some", "main", "several", "various"
- Vague catch-all: "detailed info", "personal info etc." -> pick at most 4 specific fields
- Result count numbers: "49 lawmakers", "191 minutes" -> the user does not know the count
- Generalizing domain terms: "expressway traffic" -> "traffic info" (X), "traffic" (X)
- Placeholder substitution: "specific city", "the relevant region", "specific route", "specific organization", "specific year" -- if a concrete value is in the arguments, it MUST be used. If the code cannot be naturalized into Korean, keep the code value (e.g., "city code 41150")

## Output
JSON:
{
  "sub_query": "natural Korean sub-question",
  "core_intent": "core intent (2-5 words)"
}
\end{lstlisting}
\end{tcolorbox}
\captionof{table}{Prompt for generating a natural-language sub-query for a single
tool-call step in Phase~B.}
\label{tab:prompt-phaseb-subquery}
\twocolumn

\onecolumn
\begin{tcolorbox}[
  breakable,
  colback=white,
  colframe=black,
  colbacktitle=gray!20,
  coltitle=black,
  title={\textit{\textbf{Sub-Query Composition}}},
  fonttitle=\normalsize,
  boxrule=0.4pt,
  left=3pt, right=3pt, top=2pt, bottom=2pt,
  halign title=center
]
\begin{lstlisting}[basicstyle=\ttfamily\scriptsize, breaklines=true, columns=fullflexible, keepspaces=true, language={}, showstringspaces=false, frame=none]
You compose multiple sub-queries into ONE natural Korean question.

## Input
- Sub-queries: {sub_queries_str}
- Junction Info: {junction_info_str}

## Steps (follow in order)
### Step 1: Apply pattern-specific composition rule
{pattern_rule}
### Step 2: Determine answer type
{answer_type_rule}
### Step 3: Write the composed query
Merge sub-queries into one question following the pattern rule and answer type. Preserve exact domain terminology from sub-queries
(e.g., "expressway traffic" stays as-is, do NOT generalize to "traffic info").

## FORBIDDEN (any of these -> invalid output)
- Procedural language: "retrieved", "searched", "verified", "called", "fetched", "first", "then"
- Data format / schema mentions: "JSON", "json", "format", "schema", "field" -- the user does not need to know the backend data format
- Vague terms: "recent", "some", "main", "latest"
- API field names: PRDC_YM_NM, ROUTE_CD, ORG_CD, STN_NM, etc.
- Result count numbers: "out of 49 lawmakers", "191 minutes" -- user does not know counts
- Vague catch-all: "detailed info", "personal info etc." -> max 4 specific fields
- Enumeration: "list all", "enumerate everything" -> ask count instead ("total how many?")
- Generalizing domain terms: "real-time traffic" -> "traffic info" (X)
- Intermediate result leakage (connected fields only): if step N produces a value that becomes step N+1's input argument (connected field), do NOT use that value in the query -- it enables shortcutting.
Result fields that are NOT passed to the next step may appear in the query.
- Placeholder substitution: "specific city", "the relevant region", "specific route", "specific organization", "specific year" -- if the sub-query has a concrete value it MUST be reflected. NEVER use "specific X"
- **Invented names / synthesized identifiers**: if the sub-query contains code values (NAAS_CD=14M56632, CURR_COMMITTEE_ID=9700407, etc.), keep them **as-is**. Do NOT invent person names (e.g., Hong Gildong, Kim Cheolsu), committee names (e.g., National Assembly Library, etc.). If you do not know which name corresponds to the code, write it as "code + description" e.g., "the lawmaker with code 14M56632").

## INPUT VALUE NATURALIZATION (apply whenever an input value is used in the query)
- Term/session codes -> natural language: "21" -> "21st", "414" -> "414th session"
- Date codes -> natural language: "2023-04" -> "April 2023", "20230401" -> "April 1, 2023"
- Year -> natural language: "2023" -> "year 2023"
- Quarter -> natural language: "1" -> "Q1", "4" -> "Q4"
- Code values may stay as-is, but pair with a Korean description when meaningful: "committee code AO" (OK), "city code 41390" (OK)

## MINIMALITY RULE (every step must be necessary)
Each step must feed data to the next step -- the composed query must make EVERY step logically necessary.
- The query must describe WHY the first step is needed (what it provides to the next step).
- If removing step 1 would still allow answering the query -> the query is BAD.
- BAD:  "Based on the term of the 21st National Budget Committee's minutes, how many legislative activities are there in total?" (term=21 is already in the query -> step 1 is skippable)
- GOOD: "For the term in which the Budget Committee held the most meetings, how many legislative activities are there in total?"
(step 1 is needed to find which term)

## Output
JSON:
{
  "query": "natural Korean composed question",
  "step_justifications": [
    {"step": 1, "reason": "why this step is needed"},
    {"step": 2, "reason": "why this step is needed"}
  ]
}
\end{lstlisting}
\end{tcolorbox}
\captionof{table}{Prompt that composes multiple sub-queries into a single natural
Korean question in Phase~B, enforcing that every step is logically
necessary.}
\label{tab:prompt-phaseb-compose}
\twocolumn

\onecolumn

\begin{tcolorbox}[
  breakable,
  colback=white,
  colframe=black,
  colbacktitle=gray!20,
  coltitle=black,
  title={\textit{\textbf{Answer Generation}}},
  fonttitle=\normalsize,
  boxrule=0.4pt,
  left=3pt, right=3pt, top=2pt, bottom=2pt,
  halign title=center
]
\begin{lstlisting}[basicstyle=\ttfamily\scriptsize, breaklines=true, columns=fullflexible, keepspaces=true, language={}, showstringspaces=false, frame=none]
You generate a concrete answer for the given query based on execution results.

## Query
{query}

## Execution Results (full step trace, for context)
{execution_results_str}
{answer_material_block}

## Rules
- Answer MUST contain concrete values (numbers, names, dates) from the actual results.
- **COUNT questions**: the number in the answer MUST be **taken verbatim from the `count` value of the terminal_calls material**.
  - Do NOT count records/rows yourself. Even when rows is a sample, the real total is `count`.
  - e.g., if count=5, the answer is "5 cases" (NOT 3 cases or 0 cases).
- **Distinguishing parallel calls (CMP / SEM / FAN)**: when there are multiple entries in `terminal_calls`, **match each call's input_args with its count/rows by the input values** in the answer.
  - e.g., [{"input_args": {"SIGUN_CD": "41280"}, "count": 5},
           {"input_args": {"SIGUN_CD": "41190"}, "count": 5}]
    -> answer "City code 41280: 5 cases, 41190: 5 cases".
  - None of them must be omitted.
- **VALUE questions**: use only the actual values present in rows. Do NOT invent names/values that are not in rows.
- **When there are two or more targets, separate each target on its own bullet/numbered line** (for easier verification):
  Example (CMP / SEM parallel calls):
  - City code 41280: 5 cases
  - City code 41190: 5 cases
  Example (multiple organizations):
  - National Assembly Library: 422 cases
  - National Assembly Budget Office: 132 cases
  Each bullet must lead with the target identifier. A plain sentence is allowed when there is only one target.
- **COND-pattern-only structure (queries with conditional branching)**:
  1. **You MUST check the `MATCHED BRANCH` marker in the execution results**. seed count=0 means the "none (else)" branch.
  2. **Write as two separate sentences**: the first states only the condition outcome (ending with a period); the second states only the matched branch's result.
     - seed count > 0: "[condition] exists. [matched branch result]"
     - seed count == 0: "[condition] does not exist. [matched branch result]"
  3. Use only the result values of the matched branch in the answer.
     Data for the unmatched branch is not in the trace, so do NOT include it in the answer.
  - Example (seed=0, else): "There is no lawmaker with code HE428991.
    The number of National Assembly committee status items matching the committee name '2002 World Cup International Sports Event Support Special' is 3 in total."
  - Example (seed>0, then): "There are game producers in Gyeonggi-do. There are 9 facilities available for soccer."
  - FORBIDDEN: joining the two clauses with causal connectives such as "because none exists" or "because it exists"
  - FORBIDDEN: listing values directly without the branch-selection sentence
  - FORBIDDEN: an answer that starts with "exists" when seed count=0
- FORBIDDEN placeholders: "a list is provided", "can be verified", "verifiable from the data"
- If result is empty (count=0): "There is no data matching this condition"
- Answer should directly and specifically answer the query.
- Keep the answer concise but complete.

## Output
JSON:
{
  "answer_reasoning": "step-by-step explanation of how the answer was
                      derived (which tool's which result was used; how
                      counts/values were combined)",
  "answer": "concrete answer",
  "answer_fields": ["field1", "field2"]
}
- answer_reasoning: the reasoning trace that derived the answer. Describe which execution step's results were used as evidence.
- answer_fields: list of **field names** in the result records from which the answer's actual values were taken.
  e.g., if you answered with a stock name, ["jmNm"]; with name and grade, ["jmNm", "grdNm"]. When only a count is answered, [] (empty array).
\end{lstlisting}
\end{tcolorbox}
\captionof{table}{Prompt that generates the final answer for a query from the
execution-result trace in Phase~B.}
\label{tab:prompt-phaseb-answer}
\twocolumn

\onecolumn
\begin{tcolorbox}[
  breakable,
  colback=white,
  colframe=black,
  colbacktitle=gray!20,
  coltitle=black,
  title={\textit{\textbf{Answer Leakage Detection}}},
  fonttitle=\normalsize,
  boxrule=0.4pt,
  left=3pt, right=3pt, top=2pt, bottom=2pt,
  halign title=center
]
\begin{lstlisting}[basicstyle=\ttfamily\scriptsize, breaklines=true, columns=fullflexible, keepspaces=true, language={}, showstringspaces=false, frame=none]
You are a classifier evaluating the data quality of queries used for testing Korean tool-using agents.

Target Class (answer_leaked_conditional): Cases where a query contains a conditional/filter clause meant to determine the target object, but the value of that target object is already explicitly stated elsewhere in the query, rendering the conditional check vacuous (logically meaningless).

Core Rubrics:
- Does the query contain a conditional clause such as "Z of [Object] that has both X and Y" or "Z of [Object] satisfying X and Y"?
- [Important-A] Is the actual value of the [Object] (e.g., municipality name, department name, route name) already specified elsewhere in the query text? Values embedded within actions (tool calls) do NOT constitute a leak---this can be a normal sequential operation where the agent discovers and uses the value from a prior step's result.
- [Important-B] Does the entity type of the suspected leaking value match that of the [Object] to be determined by the conditional clause? If the conditional clause determines a municipality but the query explicitly mentions a different attribute (e.g., phone number, address), it is not leakage. The leaked value must match the target entity type of the conditional clause for the check to be vacuous.
- [Important-C] If the intermediate clause's result is explicitly stated in RESPONSE_VALUE, that clause is a required answer component for evaluation---hence, not leakage (likely a parallel call or essential info).
- Is the conditional check merely formal because the [Object] is already determined (solely by looking at the query) without going through the conditions?

True Cases (remove=true):
- "In Suwon City, what is the Z of the municipality that has both X and Y?" -> The "municipality with both X and Y" should be determined via filtering, but "Suwon City" is already explicitly mentioned in the query.
- "Which facility in Icheon City has both a pottery workshop and a Confucian school with an area >= 4000?" -> The target of the conditional check is already explicitly stated as "Icheon City", making the condition vacuous.

False Cases (remove=false):
- Simple single lookup ("What is the population of Suwon City?")
- Valid conditional/comparison query---The conditional/comparison result actually determines the target of the answer:
  * "Party names of the National Assembly that has fewer minutes between the 21st and 14th assemblies" -> It must be determined through the check which assembly is referred to.
  * "Population of the municipality with the largest number of libraries" -> The specific municipality must be determined through the conditional check.
- Valid lookup query---Simple argument usage without conditional clauses:
  * "How many charter bus companies are there in Suwon City?" -> No conditional clause at all, just a lookup.
- Valid sequential discovery---The conditional clause is in the query, the [Object] value is not in the query, and actions discover the [Object] from a prior tool result to use in the next step:
  * "Status of housing land development projects in municipalities with a permitted quarrying area <= 5688 among permitted quarrying companies?" -> No municipality name in the query. actions filter step 1 results to derive the municipality, then use it in step 2. This is a valid multi-step query where the condition actually determines the answer.
  * Do not judge it as a leak just by seeing the municipality name (e.g., "Icheon City") embedded in the actions.

- When the suspected leaking value and the conditional clause target entity are of different types:
  * "At company X, what is the phone number of...?" (The company name is specified, and the phone number is asked) -> The company name is specified, but the attribute being asked is the phone number. Unless the conditional clause determines the company name, it is not leaked.
- When the intermediate result is included in RESPONSE_VALUE---that clause is essential information for evaluation:
  * If the query says "Check A and what about B?" and response.value also includes the answer to A -> Clause A is not leakage (it is either a parallel call or part of the final answer).

The reference RESPONSE_VALUE (GT) and ACTIONS are provided together. Adhere to the following principles:
- The primary baseline for determining answer leakage is the query text---actions serve only as a secondary signal.
- If a value embedded within actions is also explicitly specified in the query -> Strong signal of answer leakage.
- If a value embedded within actions is absent from the query -> Presumed to be a sequential discovery, not answer leakage.
- If the answer to an intermediate clause is included in RESPONSE_VALUE -> That clause is required for evaluation, not leakage.

Output must strictly follow the JSON schema below (No markdown, comments, or extra text):
{
  "remove": true | false,
  "explanation": "one-sentence reason"
}
\end{lstlisting}
\end{tcolorbox}
\captionof{table}{Prompt utilized in Data Filtering to detect and filter out answer-leaking queries.}
\label{tab:prompt-data-filtering}
\twocolumn

\onecolumn
\begin{tcolorbox}[
  breakable,
  colback=white,
  colframe=black,
  colbacktitle=gray!20,
  coltitle=black,
  title={\textit{\textbf{Prompt for the Agent under Test}}},
  fonttitle=\normalsize,
  boxrule=0.4pt,
  left=3pt, right=3pt, top=2pt, bottom=2pt,
  halign title=center
]
\begin{lstlisting}[basicstyle=\ttfamily\scriptsize, breaklines=true, columns=fullflexible, keepspaces=true, language={}, showstringspaces=false, frame=none]
[BASE INSTRUCTION]
You are a Tool-Executing Agent.
Your goal is to answer the user's question accurately and efficiently using the available tools.

[RUNTIME-CONTEXT INJECTION -- appended only when the task defines a system state]
Use the information in <RUNTIME_CONTEXT> when it is directly needed to execute the tool calls.
<RUNTIME_CONTEXT>
{key_1}: {value_1}
{key_2}: {value_2}
...
</RUNTIME_CONTEXT>

[RESPONSE-FORMAT SUFFIX -- selected by the task's response type]

(String)
At the very end, output your final answer in exactly this format:
ANSWER: {}
The answer must appear inside the curly braces and nowhere else.

(Number)
Put your final answer within \boxed{} as a sympy-parseable number or expression only (e.g., \boxed{42}, \boxed{sqrt(2)}).
No text or explanations - just the numerical value or mathematical expression.

(Judge)
At the very end, output your final answer in exactly this format:
ANSWER: {}
Place your comprehensive and explicit answer inside the braces, fully addressing all aspects of the question.
\end{lstlisting}
\end{tcolorbox}
\captionof{table}{System prompt shown to the agent under test, assembled per task from a base instruction, an optional runtime-context block, and a response-format suffix selected by the task's response type.}
\label{tab:prompt-agent}
\twocolumn

\onecolumn
\begin{tcolorbox}[
  breakable,
  colback=white,
  colframe=black,
  colbacktitle=gray!20,
  coltitle=black,
  title={\textit{\textbf{LLM-as-a-Judge Prompt (String / Judge)}}},
  fonttitle=\normalsize,
  boxrule=0.4pt,
  left=3pt, right=3pt, top=2pt, bottom=2pt,
  halign title=center
]
\begin{lstlisting}[basicstyle=\ttfamily\scriptsize, breaklines=true, columns=fullflexible, keepspaces=true, language={}, showstringspaces=false, frame=none]
Your task is to verify whether a student's answer matches the correct string answer. You are NOT expected to derive or infer a new correct answer - your only job is to determine whether the student's final answer semantically matches the provided correct answer.

Grading Guidelines:
- The student answer must express the same meaning as the correct answer.
- Minor grammatical or stylistic differences are allowed.
- However, if grammar is so broken that the meaning becomes unclear, confusing, or significantly degraded, REJECT.
- Synonyms and paraphrases are acceptable as long as they preserve meaning.
- Additional irrelevant information is acceptable only if it does not contradict or obscure the core meaning.
- If the student's answer contains additional information beyond the correct
  answer: ACCEPT if it does not distort the overall meaning and introduces no logical contradiction.
- If the student's response contains mutually contradictory statements or conclusions, REJECT.
- If the student's response consists of or contains meaningless repetition of tokens, words, or phrases, REJECT.

[Student Question, Student Answer, and Correct Answer inserted here.]

Go step by step through the grading criteria. Do not draw any conclusions until you have finished your reasoning.
End your response with exactly one of the following two lines (no markdown, no bold, no extra punctuation):
Decision: ACCEPT
Decision: REJECT
\end{lstlisting}
\end{tcolorbox}
\captionof{table}{LLM-as-a-Judge prompt shared by the \textit{string} and \textit{judge} response types, used as a fallback when deterministic extraction and matching fail.}
\label{tab:prompt-judge-string}
\twocolumn

\onecolumn
\begin{tcolorbox}[
  breakable,
  colback=white,
  colframe=black,
  colbacktitle=gray!20,
  coltitle=black,
  title={\textit{\textbf{LLM-as-a-Judge Prompt (Number)}}},
  fonttitle=\normalsize,
  boxrule=0.4pt,
  left=3pt, right=3pt, top=2pt, bottom=2pt,
  halign title=center
]
\begin{lstlisting}[basicstyle=\ttfamily\scriptsize, breaklines=true, columns=fullflexible, keepspaces=true, language={}, showstringspaces=false, frame=none]
Your task is to verify whether a student's answer to a specific question is correct, based on a provided correct answer. You are not expected to solve the question yourself - only to evaluate whether the student's final answer matches the correct one.

Grading Guidelines:
- Only the final answer should be evaluated. Disregard mistakes made during intermediate steps.
- Ignore formatting issues such as LaTeX errors, but do not ignore mathematically significant notation errors.
- Any discrepancy that alters the mathematical meaning of the answer should result in rejection.
- If the student's answer is a computable expression (e.g., "1 + 2 + 3", "sqrt(16) + 1"), compute it first, then compare. Do not reject solely because it is not in simplified form.
- Final answers may be unsimplified or in equivalent forms.
- Missing units can be ignored.
- When comparing decimals, use the shorter decimal as the precision standard; if the longer one rounds to match, ACCEPT (e.g., correct 3.14 vs. student 3.14159 -> ACCEPT).
- For non-math questions, the answer need not mirror the correct answer in all details, as long as it reasonably and completely addresses the core aspects.
- If the response contains mutually contradictory statements, REJECT.
- If the response consists of or contains meaningless repetition of tokens, words, or phrases, REJECT.

[Student Question, Student Answer, and Correct Answer inserted here.]

Go step by step through the grading criteria. Do not draw any conclusions until you have finished your reasoning.
End your response with exactly one of the following two lines (no markdown, no bold, no extra punctuation):
Decision: ACCEPT
Decision: REJECT
\end{lstlisting}
\end{tcolorbox}
\captionof{table}{LLM-as-a-Judge prompt for the \textit{number} response type, extending the grading guidelines with numeric-equivalence rules.}
\label{tab:prompt-judge-number}
\twocolumn

\end{document}